\documentclass[10pt,twocolumn,letterpaper]{article}
\usepackage[accsupp]{axessibility}
\usepackage{wacv}

\usepackage{times}
\usepackage{epsfig}
\usepackage{graphicx}
\usepackage{amsmath}
\usepackage{amssymb}
\usepackage{booktabs}

\usepackage{comment}
\usepackage{amsmath,amssymb}

\usepackage{color}
\usepackage{xspace}
\usepackage{bbm}

\usepackage{epsfig}
\usepackage{xcolor,colortbl}
\usepackage{booktabs}
\usepackage{pifont}
\usepackage{subfig}
\usepackage{subfiles}
\usepackage{csquotes}
\usepackage{multirow}
\usepackage{wrapfig}
\usepackage{adjustbox}

\usepackage{makecell}
\usepackage{listings}
\usepackage{xcolor}
\usepackage{algorithm}
\usepackage{algorithmic}
\usepackage{url}

\def\ours{\texttt{\textbf{SLM}}\xspace}
\definecolor{amethyst}{rgb}{0.6, 0.4, 0.8}

\wacvfinalcopy

\ifwacvfinal
\usepackage[breaklinks=true,bookmarks=false]{hyperref}
\else
\usepackage[pagebackref=true,breaklinks=true,colorlinks,bookmarks=false]{hyperref}
\fi

\begin{document}

\title{Select, Label, and Mix: Learning Discriminative Invariant Feature Representations for Partial Domain Adaptation}

\author{Aadarsh Sahoo$^{1}$ \ \ \ \ Rameswar Panda$^{1}$ \ \ \ \ Rogerio Feris$^{1}$ \ \ \ \ Kate Saenko$^{1,2}$ \ \ \ \ Abir Das$^{3}$ \\
$^{1}$ MIT-IBM Watson AI Lab, $^{2}$ Boston University, $^{3}$ IIT Kharagpur
\\
{\tt \small \{aadarsh@, rpanda@, rsferis@us.\}ibm.com},
{\tt \small saenko@bu.edu}, {\tt \small abir@cse.iitkgp.ac.in}
} 

\maketitle
\thispagestyle{empty}

\begin{abstract}
Partial domain adaptation which assumes that the unknown target label space is a subset of the source label space has attracted much attention in computer vision. Despite recent progress, existing methods often suffer from three key problems: negative transfer, lack of discriminability, and domain invariance in the latent space. To alleviate the above issues, we develop a novel `Select, Label, and Mix' (SLM) framework that aims to learn discriminative invariant feature representations for partial domain adaptation. 
First, we present an efficient \enquote{select} module that automatically filters out the outlier source samples to avoid negative transfer while aligning distributions across both domains. Second, the \enquote{label} module iteratively trains the classifier using both the labeled source domain data and the generated pseudo-labels for the target domain to enhance the discriminability of the latent space.
Finally, the \enquote{mix} module utilizes domain mixup regularization jointly with the other two modules to explore more intrinsic structures across domains leading to a domain-invariant latent space for partial domain adaptation.
Extensive experiments on several benchmark datasets for partial domain adaptation demonstrate the superiority of our proposed framework over state-of-the-art methods. Project page: \textnormal{\url{https://cvir.github.io/projects/slm}}.
\end{abstract}

\section{Introduction}
\label{sec:introduction}

Deep neural networks usually have recently shown impressive performance on many visual tasks by leveraging large collections of labeled data. However, they usually
do not generalize well to domains that are not distributed identically to the training data.
Domain adaptation~\cite{csurka2017domain,wang2018deep} addresses 
this problem by transferring knowledge from a label-rich source domain to a target domain where labels are scarce or unavailable.
However, standard domain adaptation algorithms often assume that the source and target domains share the same label space~\cite{ganin2015unsupervised,gretton2012kernel,long2015learning,long2018conditional,long2016unsupervised}. Since large-scale labelled datasets are readily accessible as source domain data, a more realistic scenario is partial domain adaptation (PDA), which assumes that target label space is a subset of source label space, that has received increasing research attention recently~\cite{bucci2019tackling,chen2019domain,chen2020selective,hu2019multi}.

Several methods have been proposed to solve partial domain adaptation by reweighting the source domain samples~\cite{bucci2019tackling,chen2019domain,chen2020selective,hu2019multi,xu2019larger,zhang2018importance}. 
However, (1) most of the existing methods still suffer from negative transfer due to presence of outlier source domain classes, which cripples domain-wise transfer with untransferable knowledge; (2) in absence of the labels, they often neglect the class-aware information in target domain which fails to guarantee the discriminability of the latent space; and (3) given filtering of the outliers, limited number of samples from the source and target domain are not alone sufficient to learn domain invariant features for such a complex problem. As a result, a domain classifier may falsely align the unlabeled target samples with samples of a different class in the source domain, leading to inconsistent predictions. 

\begin{figure*}[!t]
	\begin{center}
	\includegraphics[width=0.95\linewidth]{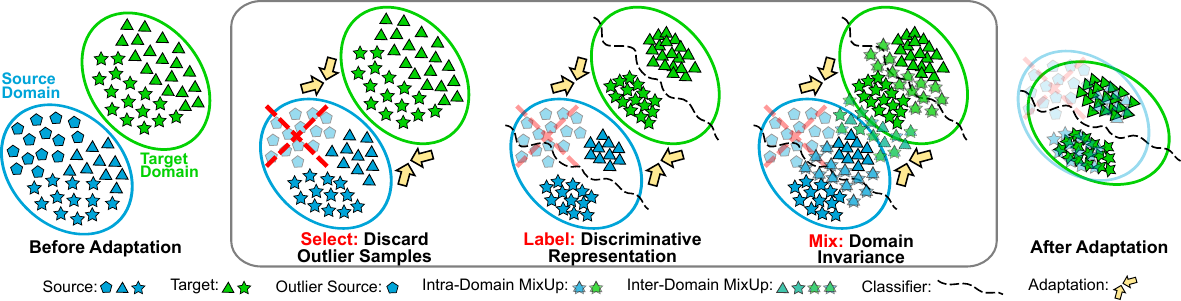}
	\end{center}
    \vspace{-4mm}
	\caption{\small \textbf{A conceptual overview of our approach}. Our proposed approach adopts three unique modules namely Select, Label and Mix in a unified framework to mitigate domain shift and 
	generalize the model to an unlabelled target domain with a label space which is a subset of that of the labelled source domain. Our Select module discards outlier samples from the source domain to eliminate negative transfer of untransferable knowledge. On the other hand, Label and Mix modules ensure discriminability and invariance of the latent space respectively while adapting the source classifier to the target domain in partial domain adaptation setting. Best viewed in color.}
	\label{fig:teaser}
    \vspace{-3mm}
\end{figure*}

To address these challenges, we propose a novel end-to-end \textbf{Select, Label, and Mix (\ours)} framework for learning discriminative invariant features while preventing negative transfer in partial domain adaptation. Our framework consists of three unique modules working in concert, \textit{i.e.}, select, label and mix, as shown in Figure~\ref{fig:teaser}. First, the select module facilitates the identification of relevant source samples preventing the negative transfer.
To be specific, our main idea is to learn a model (referred to as selector network) that outputs probabilities of binary decisions for selecting or discarding each source domain sample before aligning source and target distributions using an adversarial discriminator~\cite{ganin2016domain}. As these decision functions are discrete and non-differentiable, we rely on Gumbel Softmax sampling~\cite{jang2016categorical} to learn the policy jointly with network parameters through standard back-propagation, without resorting to complex reinforcement learning settings, as in~\cite{chen2019domain,chen2020selective}. Second, we develop an efficient self-labeling strategy that iteratively trains the classifier using both labeled source domain data and generated soft pseudo-labels for target domain to enhance the discriminabilty of the latent space. Finally, the mix module utilizes both intra- and inter-domain mixup regularizations~\cite{zhang2017mixup} to generate convex combinations of pairs of training samples and their labels in both domains. The mix strategy not only helps to explore more intrinsic structures across domains leading to an invariant latent space, but also helps to stabilize the domain discriminator while bridging distribution shift across domains.

\textit{Our proposed modules are simple yet effective which explore three unique aspects for the first time in partial domain adaptation setting in an end-to-end manner}. Specifically, in each mini-batch, our framework simultaneously eliminates negative transfer by removing outlier source samples and learns discriminative invariant features by labeling and mixing samples.
Experiments on four datasets illustrate the effectiveness of our proposed framework in achieving new state-of-the-art performance for partial domain adaptation (e.g., our approach outperforms DRCN~\cite{li2020deep} by $\mathbf{18.9\%}$ on the challenging VisDA-2017~\cite{peng2017visda} benchmark). 
To summarize, our key contributions include:

\vspace{-1mm}
\begin{itemize}
\setlength{\itemsep}{-1pt}
    \item We propose a novel Select, Label, and Mix (\ours) framework for learning discriminative and invariant feature representation while preventing intrinsic negative transfer in partial domain adaptation. 
    
    \vspace{1mm}
    
    \item We develop a simple and efficient source sample selection strategy where the selector network is jointly trained with the domain adaptation model using backpropagation through Gumbel Softmax sampling.
    
     \vspace{1mm}
     
    \item We conduct extensive experiments on four datasets, including Office31~\cite{saenko2010adapting}, Office-Home~\cite{venkateswara2017deep}, ImageNet-Caltech, and VisDA-2017~\cite{peng2017visda} to demonstrate superiority of our approach over state-of-the-art methods.
\end{itemize}

\begin{figure*}[!t]
	\begin{center}	\includegraphics[width=0.96\linewidth]{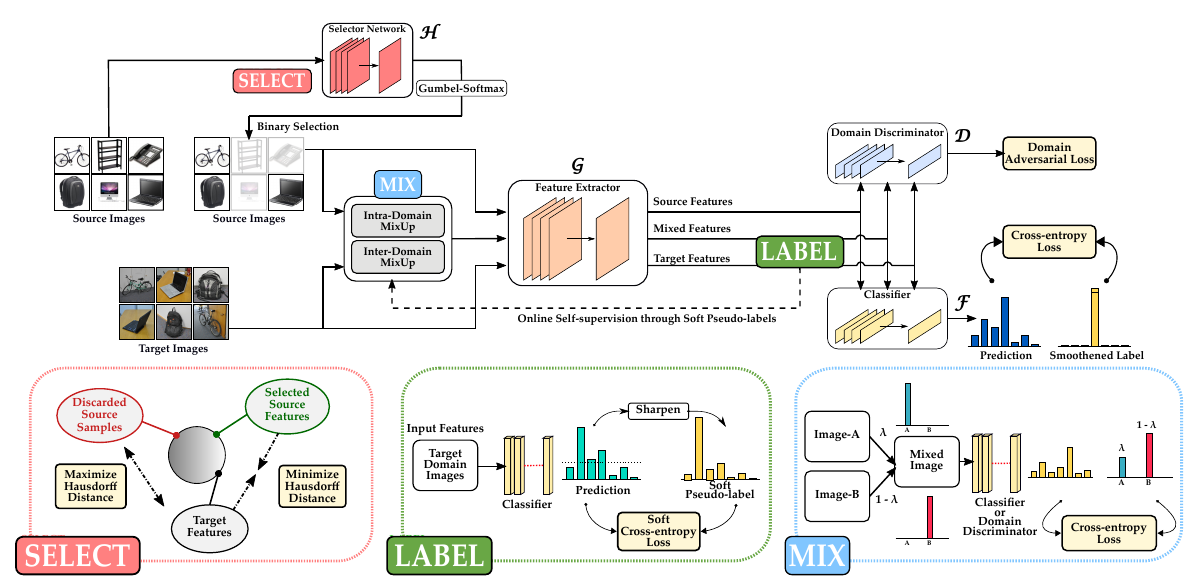}
	\end{center}
  \vspace{-4mm}
	\caption{\small \textbf{Illustration of our proposed framework}. 
	Our framework consists of a feature extractor $\mathcal{G}$ which maps the images to a common latent feature space, a classifier network $\mathcal{F}$ to provide class-wise predictions, a domain discriminator $\mathcal{D}$ to reduce domain discrepancy, and a selector network $\mathcal{H}$ for discarding outlier source samples (\enquote{Select}) to mitigate the problem of negative transfer in partial domain adaptation. Our approach also comprises of two additional modules namely \enquote{Label} and \enquote{Mix} that works in conjunction with the \enquote{Select} module to ensure the discriminability and domain invariance of the latent space. Given a mini-batch of source and target domain images, all the components are optimized jointly in an iterative manner. See Section~\ref{sec:proposedmethod} for more details. Best viewed in color.}
	\label{fig:main_figure}
	\vspace{-3mm}
\end{figure*}

\vspace{-3mm}
\section{Related Works}
\label{sec:relatedwork}
\vspace{-1mm}

\noindent\textbf{Unsupervised Domain Adaptation.} 
Unsupervised domain adaptation which aims to leverage labeled source domain data to learn to classify unlabeled target domain data has been studied from multiple perspectives (see reviews~\cite{csurka2017domain, wang2018deep}). 
Various strategies have been developed for unsupervised domain adaptation, including methods for reducing cross-domain divergence~\cite{gretton2012kernel,long2015learning,shen2017wasserstein,sun2016deep}, adding domain discriminators for adversarial training~\cite{chen2019joint,ganin2015unsupervised,ganin2016domain,long2015learning,long2018conditional,long2016unsupervised,pei2018multi,tzeng2017adversarial} and image-to-image translation techniques~\cite{hoffman2018cycada,hu2018duplex,murez2018image}.  
UDA methods assume that label spaces across source and target domains are identical unlike the practical problem we consider in this work. 

\vspace{0.5mm}
\noindent\textbf{Partial Domain Adaptation.} 
Representative PDA methods train domain discriminators~\cite{cao2018partialsan,cao2018partialpada,zhang2018importance} with weighting, or use residual correction blocks~\cite{li2020deep,liang2020balanced}, or use source examples based on their similarities to target domain~\cite{cao2019learning}. Most relevant to our approach is the work in~\cite{chen2019domain,chen2020selective} which uses Reinforcement Learning (RL) for source data selection in partial domain adaptation. RL policy gradients are
often complex, unwieldy to train and require techniques to reduce variance during training.
By contrast, our approach utilizes a gradient based optimization for relevant source sample selection which is extremely fast and computationally efficient.  
Moreover, while prior PDA methods try to reweigh source samples in some form or other, they often do not take class-aware information in target domain into consideration. Our proposed approach instead, ensures discriminability and invariance of the latent space by considering both pseudo-labeling and cross-domain mixup with sample selection in an unified framework for PDA.

\vspace{0.5mm}
\noindent\textbf{Self-Training with Pseudo-Labels.} Deep self-training methods that focus on iteratively training the model by using both labeled source data and generated target pseudo-labels have been proposed for aligning both domains~\cite{inoue2018cross,mei2020instance,saito2017asymmetric,zhang2020label}.
Majority of the methods directly choose hard pseudo-labels with high prediction confidence.
The works in~\cite{zou2019confidence,zou2018unsupervised} use class-balanced confidence regularizers to generate soft pseudo-labels for unsupervised domain adaptation that share same label space across domains. Our work on the other hand iteratively utilizes soft pseudo-labels within a batch by smoothing one-hot pseudo-label to a conservative target distribution for PDA.

\vspace{0.5mm}
\noindent\textbf{Mixup Regularization.} Mixup regularization~\cite{zhang2017mixup} or its variants~\cite{berthelot2019mixmatch,verma2019manifold} 
that train models on virtual examples constructed as convex combinations of pairs of inputs and labels are recently used to improve the generalization of neural networks. A few recent methods apply Mixup, but mainly for UDA to stabilize domain discriminator~\cite{wu2020dual,xu2019adversarial,yan2020improve} or to smoothen the predictions~\cite{mao2019virtual}. Our proposed SLM strategy can be regarded as an extension of this line of research by introducing both intra-domain and inter-domain mixup not only to stabilize the discriminator but also to guide the classifier in enriching the intrinsic structure of the latent space to solve the more challenging PDA task.

\section{Methodology}
\label{sec:proposedmethod}

Partial domain adaptation aims to mitigate the domain shift and generalize the model to an unlabelled target domain with a label space which is a subset of that of the labelled source domain. Formally, we define the set of labelled source domain samples as $\mathcal{D}_{source}\!=\!\{(\textbf{x}_{i}^{s}, y_{i})\}_{i=1}^{N_{S}}$ and unlabelled target domain samples as $\mathcal{D}_{target}\!=\!\{\textbf{x}_{i}^{t}\}_{i=1}^{N_{T}}$, with label spaces $\mathcal{L}_{source}$ and $\mathcal{L}_{target}$, respectively, where $\mathcal{L}_{source}\!\subsetneq\!\mathcal{L}_{target}$.
$N_{S}$ and $N_{T}$ represent the number of samples in source and target domain respectively.
Let $p$ and $q$ represent the probability distribution of data in source and target domain respectively. In partial domain adaptation, we further have $p\!\neq\!q$ and $p_{\mathcal{L}_{target}}\!\neq\!q$, where $p_{\mathcal{L}_{target}}$ is the distribution of source domain data in $\mathcal{L}_{target}$.
Our goal is to develop an approach with the above given data to improve the performance of a model on $\mathcal{D}_{target}$.

\vspace{-2mm}
\subsection{Approach Overview}
\vspace{-1mm}

Figure~\ref{fig:main_figure} illustrates an overview of our proposed approach. Our framework consists of a feature extractor $\mathcal{G}$, a classifier network $\mathcal{F}$, a domain discriminator $\mathcal{D}$ and a selector network $\mathcal{H}$. Our goal is to improve classification performance of the combined network $\mathcal{F(G(.))}$ on $\mathcal{D}_{target}$. While the feature extractor $\mathcal{G}$ maps the images to a common latent space, the task of classifier $\mathcal{F}$ is to output a probability distribution over the classes for a given feature from $\mathcal{G}$. Given a feature from $\mathcal{G}$, the discriminator $\mathcal{D}$ helps in minimizing domain discrepancy by identifying the domain (either source or target) to which it belongs. The selector network $\mathcal{H}$ helps in reducing negative transfer by learning to identify outlier source samples from $\mathcal{D}_{source}$ using Gumbel-Softmax sampling~\cite{jang2016categorical}.
On the other hand, label module utilizes predictions of $\mathcal{F(G(.))}$ to obtain soft pseudo-labels for target samples. Finally, the mix module leverages both pseudo-labeled target samples and source samples to generate augmented images for achieving domain invariance in the latent space. During training, for a mini-batch of images, all the components are trained jointly and during testing, we evaluate performance using classification accuracy of the network $\mathcal{F(G(.))}$ on target domain data $\mathcal{D}_{target}$. The individual modules are discussed below.

\subsection{Select Module}

\begin{figure}[!t]
	\begin{center}
	\includegraphics[width=0.45\textwidth]{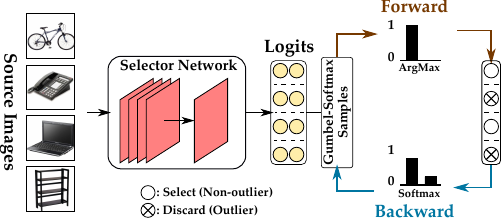} 
	\end{center}
    \vspace{-3mm}
	\caption{\small \textbf{Learning with Gumbel Softmax Sampling.} Gumbel-Softmax trick used for enabling gradient-based optimization for discrete output space. Best viewed in color.}
	\label{fig:gumbel_softmax}
	\vspace{-5mm}
\end{figure}

This module stands in the core of our framework with an aim to get rid of the outlier source samples in the source domain in order to minimize negative transfer in partial domain adaptation.
Instead of using different heuristically designed criteria for weighting source samples, we develop a novel selector network $\mathcal{H}$, that takes images from the source domain as input and makes instance-level binary predictions to obtain \textit{relevant} source samples for adaptation, as shown in Figure~\ref{fig:gumbel_softmax}.
Specifically, the selector network $\mathcal{H}$ performs robust selection by providing a discrete binary output of either a 0 (\textit{discard}) or 1 (\textit{select}) for each source sample, \textit{i.e.}, $\mathcal{H}: \mathcal{D}_{source}\!\rightarrow\!\{0, 1\}$. We leverage Gumbel-Softmax operation to design the learning protocol of the selector network, as described next.
Given the selection, we forward only the selected samples to the successive modules.

\vspace{1mm}
\noindent\textbf{Training using Gumbel-Softmax Sampling.}
Our select module makes decisions about whether a source sample belongs to an outlier class or not. However, the fact that the decision policy is discrete makes the network non-differentiable and therefore difficult to optimize via standard backpropagation. To resolve the non-differentiability and enable gradient descent optimization for the selector in an efficient way, we adopt Gumbel-Softmax trick~\cite{jang2016categorical,maddison2016concrete} and draw samples from a categorical distribution parameterized by $\alpha_{0}$, $\alpha_{1}$, where $\alpha_{0}$, $\alpha_{1}$ are the output logits of the selector network for a sample to be selected and discarded respectively. As shown in Figure~\ref{fig:gumbel_softmax}, the selector network $\mathcal{H}$ takes a batch of source images (say, $\mathcal{B}_{s}$ of size $b$) as input, and outputs a two-dimensional matrix $\beta \in \mathbb{R}^{b\times2}$, where each row corresponds to [$\alpha_{0}$, $\alpha_{1}$] for an image. We then draw i.i.d. samples $G_{0}$, $G_{1}$ from $Gumbel(0, 1) = -log(-log(U))$, where $U\sim\text{Uniform}[0,1]$ and generate discrete samples in the forward pass as: $X = \arg\max_{i} [\log \alpha_i + G_i]$ resulting in hard binary predictions, while during backward pass, we approximate gradients using continuous softmax relaxation as:
\begin{align}
    \mathcal{Y}_{i} = \frac{\exp((\log\alpha_{i}\! +\! G_{i})/\tau)}{\sum_{j\in\{0, 1\}} \exp((\log\alpha_{j}\! +\! G_{j})/\tau)} \quad \textnormal{for } i\! \in\!\{0, 1\}
    \label{eq:gumbel_softmax}    
\end{align}
where $G_{i}$'s are i.i.d samples from standard Gumbel distribution $Gumbel(0, 1)$ and $\tau$ denotes temperature of softmax. When $\tau>$ 0, the Gumbel-Softmax distribution is smooth and hence gradients can be computed with respect to  logits $\alpha_{i}$'s to train the selector network using backpropagation. As $\tau$ approaches 0, $\mathcal{Y}_{i}$ becomes one-hot and discrete.

\vspace{1mm}
\noindent\textbf{Learning to Discard the Outlier Distribution.} With the unsupervised nature of this decision problem, the design of the loss function for the selector network is challenging. We propose a novel Hausdorff distance-based triplet loss function for the select module which ensures that the selector network learns to distinguish between the outlier and the non-outlier distribution in the source domain. For a given batch of source domain images $\mathcal{D}_{source}^{b}$ and target domain images $\mathcal{D}_{target}^{b}$, each of size $b$, the selector results in two subsets of source samples $\mathcal{D}_{sel}\! =\! \{ x\! \in\! \mathcal{D}_{source}^{b}\! :\! \mathcal{H}(x)\! =\! 1\}$ and $\mathcal{D}_{dis}\! =\! \{ x\! \in\! \mathcal{D}_{source}^{b}\! :\! \mathcal{H}(x)\! =\! 0\}$. The idea is to pull the \textit{selected} source samples $\mathcal{D}_{sel}$ \& target samples $\mathcal{D}_{target}^{b}$ closer while pushing \textit{discarded} source samples $\mathcal{D}_{dis}$ \& $\mathcal{D}_{target}^{b}$ apart in the output latent feature space of $\mathcal{G}$. To achieve this, we formulate the loss function as follows:
\begin{align}
    d_{sel} &= d_{H}(\mathcal{G}(\mathcal{D}_{sel}), \mathcal{G}(\mathcal{D}_{target}^{b})) \nonumber \\
    d_{dis} &= d_{H}(\mathcal{G}(\mathcal{D}_{dis}), \mathcal{G}(\mathcal{D}_{target}^{b})) \nonumber \\
    \mathcal{L}_{select} = \lambda_{s}&max(d_{sel} - d_{dis} + margin, 0) + \mathcal{L}_{reg}
    \label{eq:loss_select} 
\end{align}
where $d_{H}(X, Y)$ represents the average Hausdorff distance between the set of features $X$ and $Y$. $\mathcal{L}_{reg} = \lambda_{reg1}\sum_{x\in\mathcal{D}_{source}^{b}} \mathcal{H}(x)\log(\mathcal{H}(x)) + \lambda_{reg2}\{\sum_{\hat{p}}l_{ent}(\hat{p}) - l_{ent}(\hat{p}_{m})\}$, with $l_{ent}$ being the entropy loss, $\hat{p}$ is the Softmax prediction of $\mathcal{F(G(D}_{target}))$ and $\hat{p}_{m}$ is mean prediction for the target domain. $\mathcal{L}_{reg}$ is a regularization to restrict $\mathcal{H}$ from producing trivial all-0 or all-1 outputs as well as ensuring confident and diverse predictions by $\mathcal{F(G(.))}$ for $\mathcal{D}_{target}$. Note that only $\mathcal{D}_{sel}\!$ is used to train the classifier, domain discriminator and is utilised by other modules to perform subsequent operations.
Furthermore, to avoid any interference from the backbone feature extractor $\mathcal{G}$, we use a separate feature extractor for the select module, while making these decisions.
In summary, the supervision signal for selector module comes from (a) the discriminator directly, (b) through interactions with other modules via joint learning, and (c) the triplet loss using Hausdorff distance.

\subsection{Label Module}

While our select module helps in removing source domain outliers, it fails to guarantee the discriminability of the latent space due to the absence of class-aware information in the target domain. Specifically, given our main objective is to improve the classification performance on target domain samples, it becomes essential for the classifier to learn confident decision boundaries in the target domain. To this end, we propose a label module that provides additional self-supervision for target domain samples. Motivated by the effectiveness of confidence guided self-training~\cite{zou2019confidence}, we generate soft pseudo-labels for the target domain samples that efficiently attenuates the unwanted deviations caused by false and noisy one-hot pseudo-labels. For a target domain sample $\textbf{x}_{k}^{t} \in \mathcal{D}_{target}$, the soft-pseudo-label $\hat{y}_{k}^{t}$ is computed as follows:
\begin{align} 
    \hat{y}_{k}^{t(i)} = \frac{p(i|\textbf{x}_{i}^{t})^{\frac{1}{\alpha}}}{\sum_{j=1}^{|\mathcal{L}_{source}|}p(j|\textbf{x}_{i}^{t})^{\frac{1}{\alpha}}}
    \label{eq:soft_pseudo_label}  
\end{align}
where $p(j|\textbf{x}_{i}^{t}) = \mathcal{F(G(}\textbf{x}_{i}^{t}))^{(j)}$ is the softmax probability of the classifier for class $j$ given $\textbf{x}_{i}^{t}$ as input, and $\alpha$ is a hyper-parameter that controls the softness of the label. 
The soft pseudo-label $\hat{y}_{i}^{t}$ is then used to compute the loss $\mathcal{L}_{label}$ for a given batch of target samples $\mathcal{D}_{target}^{b}$ as follows:
\begin{align}
    \mathcal{L}_{label} = \mathbb{E}_{\textbf{x}_{i}^{t} \in \mathcal{D}_{target}^{b}} l_{ce}(\mathcal{F(G(}\textbf{x}_{i}^{t})), \hat{y}_{i}^{t})
    \label{eq:loss_label}
\end{align}
where $l_{ce}(.)$ represents the cross-entropy loss.

\subsection{Mix Module}
Learning a domain-invariant latent space is crucial for effective adaptation of a classifier from source to target domain. However, with limited samples per batch and after discarding the outlier samples, it becomes even more challenging in preventing over-fitting and learning domain invariant representation. To mitigate this problem, we apply MixUp~\cite{zhang2017mixup} on the selected source samples and the target samples for discovering ingrained structures in establishing domain invariance. Given $\mathcal{D}_{sel}$ from select module and $\mathcal{D}_{target}^{b}$ with corresponding labels $\hat{y}^{t}$ from label module, we perform convex combinations of images belonging to these two sets on pixel-level in three different ways namely, \textit{inter-domain}, \textit{intra-source} domain and \textit{intra-target} domain to obtain the following sets of augmented data:
\begin{align}
    \mathcal{D}_{\textnormal{\textit{inter\_mix}}}^{b} &= \{(\lambda\textbf{x}_{i}^{s} + (1\!-\!\lambda)\textbf{x}_{j}^{t}, \lambda y_{i} + (1\!-\!\lambda)\hat{y}_{j}^{t})\} \nonumber\\
    \mathcal{D}_{\textnormal{\textit{intra\_mix\_s}}}^{b} &= \{(\lambda\textbf{x}_{i}^{s} + (1\!-\!\lambda)\textbf{x}_{j}^{s}, \lambda y_{i} + (1\!-\!\lambda)y_{j})\} \nonumber\\
    \mathcal{D}_{\textnormal{\textit{intra\_mix\_t}}}^{b} &= \{(\lambda\textbf{x}_{i}^{t} + (1\!-\!\lambda)\textbf{x}_{j}^{t}, \lambda\hat{y}_{i}^{t} + (1\!-\!\lambda)\hat{y}_{j}^{t})\} \nonumber\\
    \mathcal{D}_{\textnormal{\textit{mix}}}^{b} &= \mathcal{D}_{\textnormal{\textit{inter\_mix}}}^{b} \cup \mathcal{D}_{\textnormal{\textit{intra\_mix\_s}}}^{b} \cup \mathcal{D}_{\textnormal{\textit{intra\_mix\_t}}}^{b}
    \label{eq:mixed_data}
\end{align}
where $(\textbf{x}_{i/j}^{s}, y_{i/j}) \in \mathcal{D}_{sel}$, while $\textbf{x}_{i/j}^{t} \in \mathcal{D}_{target}^{b}$ with  $\hat{y}_{i/j}^{t}$ being the corresponding soft-pseudo-labels. $\lambda$ is the mix-ratio randomly sampled from a beta distribution $Beta(\alpha, \alpha)$ for $\alpha \in (0, \infty)$. We use $\alpha=2.0$ in all our experiments. 
Given the new augmented images, we utilize the new augmented images in training both the classifier $\mathcal{F}$ and the domain discriminator $\mathcal{D}$ as follows:

\begin{align}
    \mathcal{L}_{mix\_cls} &= \mathbb{E}_{(\textbf{x}_{i}, y_{i}) \in \mathcal{D}_{\textnormal{\textit{mix}}}^{b}} l_{ce}(\mathcal{F(G(}\textbf{x}_{i})), y_{i}) \nonumber\\
    \mathcal{L}_{mix\_dom} &= \mathbb{E}_{\textbf{x}_{i} \sim \mathcal{D}_{\textnormal{\textit{inter\_mix}}}^{b}} [\lambda\log(\mathcal{D(G(}\textbf{x}_{i}))) \nonumber \\ &+ (1\!-\!\lambda)\log(1\!-\!\mathcal{D(G(}\textbf{x}_{i})))] \nonumber\\
    &+ \mathbb{E}_{\textbf{x}_{i} \sim \mathcal{D}_{\textnormal{\textit{intra\_mix\_s}}}^{b}} \log(\mathcal{D(G(}\textbf{x}_{i}))) \nonumber\\
    &+ \mathbb{E}_{\textbf{x}_{i} \sim \mathcal{D}_{\textnormal{\textit{intra\_mix\_t}}}^{b}} \log(1\!-\!\mathcal{D(G(}\textbf{x}_{i}))) \nonumber\\
    \mathcal{L}_{mix} &= \mathcal{L}_{mix\_cls} + \mathcal{L}_{mix\_dom}
    \label{eq:loss_mix}
\end{align}
where $\mathcal{L}_{mix\_cls}$ and $\mathcal{L}_{mix\_dom}$ represent loss for classifier and domain discriminator respectively. 
Our mix strategy with the combined loss $\mathcal{L}_{mix}$ not only helps to explore more intrinsic structures across domains leading to an invariant latent space, but also helps to stabilize the domain discriminator while bridging the distribution shift across domains.

\vspace{1mm}
\noindent\textbf{Optimization.}
Besides the above three modules that are tailored for partial domain adaptation, we use the standard supervised loss on the labeled source data and domain adversarial loss as follows:
\begin{align}
    \mathcal{L}_{sup} &= \mathbb{E}_{(\textbf{x}_{i}, y_{i}) \in \mathcal{D}_{sel}} l_{ce}(\mathcal{F(G(}\textbf{x}_{i})), y_{i})\nonumber\\
    \mathcal{L}_{adv} &= \mathbb{E}_{\textbf{x}^{s} \sim \mathcal{D}_{sel}} w^{s}\log(\mathcal{D(G(}\textbf{x}^{s}))) \nonumber\\
    &+ \mathbb{E}_{\textbf{x}^{t} \sim \mathcal{D}_{target}^{b}} w^{t}\log(1\!-\!\mathcal{D(G(}\textbf{x}^{t})))
    \label{eq:loss_standard}
\end{align}
where $\mathcal{L}_{adv}$ is entropy-conditioned domain adversarial loss with weights  $w^{s}$ and $w^{t}$ for source and target domain respectively~\cite{long2018conditional}. The overall loss $\mathcal{L}_{total}$ is
\begin{align}
    \mathcal{L}_{total} = \mathcal{L}_{sup} + \mathcal{L}_{adv} + \mathcal{L}_{select} + \mathcal{L}_{label} + \mathcal{L}_{mix}
    \label{eq:loss_total}
\end{align}
where $\mathcal{L}_{select}$, $\mathcal{L}_{label}$, and $\mathcal{L}_{mix}$ are given by Equations (\ref{eq:loss_select}), (\ref{eq:loss_label}), and (\ref{eq:loss_mix}) respectively, where we have included the corresponding weight coefficient hyperparameters. We integrate all the modules into one framework, as shown in Figure~\ref{fig:main_figure} and train the network jointly for partial domain adaptation.

\section{Experiments}
\label{sec:experiments}
\vspace{-1mm}

We conduct extensive experiments to show that our \ours framework outperforms many competing approaches to achieve the new state-of-the-art on several PDA benchmark datasets. We also perform comprehensive ablation experiments and feature visualizations to verify the effectiveness of different components in detail.

\subsection{Experimental Setup}

\noindent\textbf{Datasets.} We evaluate our approach using four datasets under PDA setting, namely Office31~\cite{saenko2010adapting}, Office-Home~\cite{venkateswara2017deep}, ImageNet-Caltech and VisDA-2017~\cite{peng2017visda}. 
Office31 contains 4,110 images of 31 classes from three distinct domains, namely Amazon (A), Webcam (W) and DSLR (D).
Following~\cite{chen2020selective}, we select 10 classes shared by Office31 and Caltech256~\cite{griffin2007caltech} as target categories. Office-Home is a challenging dataset that contains images from four domains: Artistic images (Ar), Clipart images (Cl), Product images (Pr) and Real-World images (Rw). 
We follow~\cite{chen2020selective} to select the first 25 categories (in alphabetic order) in each domain as target classes. 
ImageNet-Caltech is a challenging dataset that consists of two subsets, ImageNet1K (I)~\cite{russakovsky2015imagenet} and Caltech256 (C)~\cite{griffin2007caltech}. While source domain contains 1,000 and 256 classes for ImageNet and Caltech respectively, each target domain contains only 84 classes that are common across both domains.
VisDA-2017 is a large-scale challenging dataset with 12 categories across 2 domains: one consists 
photo-realistic images or real images (R), and the other comprises of 
and synthetic 2D renderings of 3D models (S). We select the first 6 categories (in alphabetical order) in each of the domain as the target categories\cite{li2020deep}. More details are included in the Appendix.

\vspace{1mm}
\noindent\textbf{Baselines.} We compare our approach with several methods that fall into two main categories: (1) popular UDA methods (e.g., DANN~\cite{ganin2016domain}, CORAL~\cite{sun2016deep}) including recent methods like CAN~\cite{kang2019contrastive} and SPL~\cite{wang2020unsupervised} which have shown state-of-the-art performance on UDA setup, (2) existing partial domain adaptation methods including PADA~\cite{cao2018partialpada}, SAN~\cite{cao2018partialsan}, ETN~\cite{cao2019learning}, and DRCN~\cite{li2020deep}. We also compare with the recent state-of-the-art methods, RTNet~\cite{chen2020selective} that uses reinforcement learning for source dataset selection, and BA$^3$US~\cite{liang2020balanced} which uses source samples to augment the target domain in partial domain adaptation. 

\vspace{1mm}
\noindent\textbf{Implementation Details.} 
We use ResNet-50~\cite{he2016deep} as the backbone network for the feature extractor while we use ResNet-18 for the selector network, initialized with ImageNet~\cite{russakovsky2015imagenet} pretrained weights. 
All the weights are shared by both the source and target domain images except that of the BatchNorm layers, for which we use Domain-Specific Batch Normalization~\cite{chang2019domain}. 
In Eqn.~\ref{eq:loss_select} we set $\lambda_{s}, \lambda_{reg1}$ and $\lambda_{reg2}$ as $0.01, 10.0$ and $0.1$, respectively. 
We use a margin value of $100.0$ in all our experiments. We use gradient reversal layer (GRL) for adversarially training the discriminator. We set $\tau=1.0$ in Eqn.~\ref{eq:gumbel_softmax}, $\alpha=0.1$ in Eqn.~\ref{eq:soft_pseudo_label}, and $\lambda=0.0$ for the GRL as initial values and gradually anneal $\tau$ and $\alpha$ down to 0 while increase $\lambda$ to 1.0 during the training, as in~\cite{jang2016categorical}. 
Additionally, we use label-smoothing for all the losses for the feature extractor involving source domain images as in~\cite{liang2020we,muller2019does}, with $\epsilon\!=\!0.2$.
We use SGD for optimization with momentum=0.9 while a weight decay of 1e-3 and 5e-4 for the selector network and the other networks respectively. We use an initial learning rate of 5e-3 for the selector and the classifier, while 5e-4 for the rest of the networks and decay it following a cosine annealing strategy. 
We use a batch size of 64 for Office31 and VisDA-2017 while a batch size of 128 is used for Office-Home and ImageNet-Caltech. 
We report average classification accuracy and standard deviation over 3 random trials. 
All the codes were implemented using PyTorch~\cite{paszke2019pytorch}.
\setlength{\tabcolsep}{2pt}
\begin{table}[!tbp]

\definecolor{Gray}{gray}{0.90}
\definecolor{LightCyan}{rgb}{0.88,1,1}

\newcolumntype{a}{>{\columncolor{Gray}}c}

\scriptsize
\begin{center}
\resizebox{\columnwidth}{!}{
\begin{tabular}{ l || c  c   c   c   c   c | a  }

\hline
\multicolumn{8}{c}{\textbf{Office31}} \\
\hline
 \textbf{Method} &  \textbf{A $\rightarrow$ W} & \textbf{D $\rightarrow$ W} & \textbf{W $\rightarrow$ D} & \textbf{A $\rightarrow$ D} & \textbf{D $\rightarrow$ A} & \textbf{W $\rightarrow$ A} & \textbf{Average} \\
\hline
\hline
 ResNet-50         & $76.5_{\pm 0.3}$  & $99.2_{\pm 0.2}$  & $97.7_{\pm 0.1}$  & $87.5_{\pm 0.2}$  & $87.2_{\pm 0.1}$  & $84.1_{\pm 0.3}$  & $88.7$ \\

\hline

 DANN   & $62.8_{\pm 0.6}$  & $71.6_{\pm 0.4}$  & $65.6_{\pm 0.5}$  & $65.1_{\pm 0.7}$  & $78.9_{\pm 0.3}$  & $79.2_{\pm 0.4}$  & $70.5$ \\
 CORAL  & $52.1_{\pm 0.5}$  & $65.2_{\pm 0.2}$  & $64.1_{\pm 0.7}$  & $58.0_{\pm 0.5}$  & $73.1_{\pm 0.4}$  & $77.9_{\pm 0.3}$  & $65.1$ \\
 ADDA   & $75.7_{\pm 0.2}$  &  $95.4_{\pm 0.2}$  &  \underline{$99.9$}$_{\pm 0.1}$  &  $83.4_{\pm 0.2}$  &  $83.6_{\pm 0.1}$  &  $84.3_{\pm 0.1}$  &  $87.0$ \\
 RTN  & $75.3$  & $97.1$  & $98.3$  & $66.9$  & $85.6$  & $85.7$  & $84.8$ \\
 CDAN+E   & $80.5_{\pm 1.2}$  &  $99.0_{\pm 0.0}$  &  $98.1_{\pm 0.0}$  &  $77.1_{\pm 0.9}$  &  $93.6_{\pm 0.1}$  &  $91.7_{\pm 0.0}$  &  $90.0$ \\
 JDDA   & $73.5_{\pm 0.6}$  & $93.1_{\pm 0.3}$  & $89.3_{\pm 0.2}$  & $76.4_{\pm 0.4}$  & $77.6_{\pm 0.1}$  & $82.8_{\pm 0.2}$  & $82.1$ \\
 CAN   & $84.4_{\pm 0.0}$  & $92.0_{\pm 1.4}$  & $94.7_{\pm 1.7}$  & $84.9_{\pm 0.9}$  & $85.6_{\pm 1.0}$  & $86.4_{\pm 0.8}$  & $88.0$ \\

\hline

 PADA   & $86.3_{\pm 0.4}$  & \underline{$99.3$}$_{\pm 0.1}$  & $\mathbf{100.0}_{\pm 0.0}$   & $90.4_{\pm 0.1}$  & $91.3_{\pm 0.2}$  & $92.6_{\pm 0.1}$  & $93.3$ \\
 SAN   & $93.9_{\pm 0.5}$  &  \underline{$99.3$}$_{\pm 0.5}$  &  $99.4_{\pm 0.1}$  &  $94.3_{\pm 0.3}$  &  $94.2_{\pm 0.4}$  &  $88.7_{\pm 0.4}$  &  $95.0$ \\
 IWAN   & $89.2_{\pm 0.4}$  &  \underline{$99.3$}$_{\pm 0.3}$  &  $99.4_{\pm 0.2}$  &  $90.5_{\pm 0.4}$  &  \underline{$95.6$}$_{\pm 0.3}$  &  $94.3_{\pm 0.3}$  &  $94.7$ \\
 ETN    & $93.4_{\pm 0.3}$  & \underline{$99.3$}$_{\pm 0.1}$  & $99.2_{\pm 0.2}$  & $95.5_{\pm 0.4}$  & $95.4_{\pm 0.1}$  & $91.7_{\pm 0.2}$  & $95.8$ \\
 DRCN 	& $88.1$ & $\mathbf{100.0}$ & $\mathbf{100.0}$ & $86.0$ & \underline{$95.6$} & \underline{$95.8$} & $94.3$ \\
 RTNet  & $95.1_{\pm 0.3}$  & $\mathbf{100.0}_{\pm 0.0}$   & $\mathbf{100.0}_{\pm 0.0}$   & $97.8_{\pm 0.1}$  & $93.9_{\pm 0.1}$  & $94.1_{\pm 0.1}$  & $96.8$ \\
 RTNet$_{\text{adv}}$ & $96.2_{\pm 0.3}$  & $\mathbf{100.0}_{\pm 0.0}$   & $\mathbf{100.0}_{\pm 0.0}$   & $97.6_{\pm 0.1}$  & $92.3_{\pm 0.1}$  & $95.4_{\pm 0.1}$  & $96.9$ \\
  BA$^3$US & \underline{$99.0$}$_{\pm 0.3}$  & $\mathbf{100.0}_{\pm 0.0}$   & $98.7_{\pm 0.0}$   & $\mathbf{99.4}_{\pm 0.0}$  & $94.8_{\pm 0.1}$  & $95.0_{\pm 0.1}$  & \underline{$97.8$} \\

\hline
 \textbf{\ours (Ours)} & $\mathbf{99.8}_{\pm 0.2}$  & $\mathbf{100.0}_{\pm 0.0}$  & $99.8_{\pm 0.3}$  & \underline{$98.7$}$_{\pm 0.0}$  & $\mathbf{96.1}_{\pm 0.1}$  & $\mathbf{95.9}_{\pm 0.0}$  & $\mathbf{98.4}$   \\
\hline
\end{tabular}}
\end{center}
\vspace{-5mm}
\caption{\small \textbf{Performance on Office31.} 
Numbers show the accuracy (\%) of different methods on partial domain adaptation setting. We highlight the \textbf{best} and \underline{second best} method on each transfer task. While the upper section shows the results of some popular unsupervised domain adaptation approaches, the lower section shows results of existing partial domain adaptation methods. 
\ours achieves the best performance on 4 out of 6 transfer tasks including the best average performance among all compared methods.
}
\label{table:cls-office-31} 
\vspace{-6mm}
\end{table}

\vspace{-1mm}
\subsection{Results and Analysis}
Table~\ref{table:cls-office-31} shows the results of our proposed method and other competing approaches on the Office31 dataset. We have the following key observations.
(1) As expected, the popular UDA methods including the recent CAN~\cite{kang2019contrastive}, fail to outperform the simple no adaptation model (ResNet-50), which implies that they suffer from negative transfer due to the presence of outlier source samples in partial domain adaptation.
(2) Overall, our \ours framework outperforms all the existing PDA methods by achieving the best results on \textbf{4 out of 6} transfer tasks. Among PDA methods, BA$^3$US~\cite{liang2020balanced} is the most competitive.
However, \ours still outperforms it ($97.8$\% vs $98.4$\%) due to our two novel components working in concert with the removal of outliers: enhancing discriminability of the latent space via iterative pseudo-labeling of target domain samples and learning domain-invariance through mixup regularizations.
(3) Our approach performed remarkably well on transfer tasks where the number of source domain images is very small compared to the target domain, e.g., on D $\!\rightarrow\!$ A, \ours outperforms BA$^3$US by $\mathbf{1.3}$\textbf{\%}. This shows that our method improves generalization ability of source classifier in target domain while reducing negative transfer.

\setlength{\tabcolsep}{1pt}
\begin{table*}[!thbp]

\definecolor{Gray}{gray}{0.90}
\definecolor{LightCyan}{rgb}{0.88,1,1}

\newcolumntype{a}{>{\columncolor{Gray}}c}

\scriptsize
\begin{center}
\resizebox{\linewidth}{!}{
\begin{tabular}{ l || c c c c c c c c c c c c | a  }

\hline
\multicolumn{14}{c}{\textbf{Office-Home}} \\
\hline
 \textbf{Method} &  \textbf{Ar $\!\rightarrow\!$ Cl} & \textbf{Ar $\!\rightarrow\!$ Pr} & \textbf{Ar $\!\rightarrow\!$ Rw} & \textbf{Cl $\!\rightarrow\!$ Ar} & \textbf{Cl $\!\rightarrow\!$ Pr} & \textbf{Cl $\!\rightarrow\!$ Rw} & \textbf{Pr $\!\rightarrow\!$ Ar} & \textbf{Pr $\!\rightarrow\!$ Cl} & \textbf{Pr $\!\rightarrow\!$ Rw} & \textbf{Rw $\!\rightarrow\!$ Ar} & \textbf{Rw $\!\rightarrow\!$ Cl} & \textbf{Rw $\!\rightarrow\!$ Pr} & \textbf{Average} \\
\hline
\hline

 ResNet-50 & $47.2$ & $66.8$ & $76.9$ & $57.6$ & $58.4$ & $62.5$ & $59.4$ & $40.6$ & $75.9$ & $65.6$ & $49.1$ & $75.8$ & $61.3$ \\

\hline

DANN     & $43.2$  & $61.9$  & $72.1$  & $52.3$  & $53.5$  & $57.9$  & $47.2$  & $35.4$  & $70.1$  & $61.3$  & $37.0$  & $71.7$  & $55.3$ \\
  CORAL    & $38.2$  & $55.6$  & $65.9$  & $48.4$  & $52.5$  & $51.3$  & $48.9$  & $32.6$  & $67.1$  & $63.8$  & $35.9$  & $69.8$  & $52.5$ \\
 ADDA   &  $45.2$  &  $68.8$  &  $79.2$  &  $64.6$  &  $60.0$  &  $68.3$ &  $57.6$ &  $38.9$ &  $77.5$ &  $70.3$ &  $45.2$ &  $78.3$ &  $62.8$ \\
 RTN & $49.4$  & $64.3$  & $76.2$  & $47.6$  & $51.7$  & $57.7$  & $50.4$  & $41.5$  & $75.5$  & $70.2$  & $51.8$  & $74.8$  & $59.3$ \\
 CDAN+E &  $47.5$  &  $65.9$  &  $75.7$  &  $57.1$  &  $54.1$  &  $63.4$ &  $59.6$ &  $44.3$ &  $72.4$ &  $66.0$ &  $49.9$ &  $72.8$ &  $60.7$ \\
 JDDA     & $45.8$  & $63.9$  & $74.1$  & $51.8$  & $55.2$  & $60.3$  & $53.7$  & $38.3$  & $72.6$  & $62.5$  & $43.3$  & $71.3$  & $57.7$ \\
 SPL & $46.4$  & $70.5$  & $77.2$  & $61.0$  & $65.2$  & $73.2$  & $64.3$  & $44.7$  & $79.1$  & $69.5$  & $58.0$  & $79.8$  & $65.7$ \\
\hline

 PADA    & $53.2$  & $69.5$  & $78.6$  & $61.7$  & $62.7$  & $60.9$  & $56.4$  & $44.6$  & $79.3$  & $74.2$  & $55.1$  & $77.4$  & $64.5$ \\
 SAN   &  $44.4$  &  $68.7$  &  $74.6$  &  $67.5$  &  $65.0$  &  $77.8$ &  $59.8$ &  $44.7$ &  $80.1$ &  $72.2$ &  $50.2$ &  $78.7$ &  $65.3$ \\
 IWAN   &  $53.9$  &  $54.5$  &  $78.1$  &  $61.3$  &  $48.0$  &  $63.3$ &  $54.2$ &  $52.0$ &  $81.3$ &  $76.5$ &  $56.8$ &  $82.9$ &  $63.6$ \\
 ETN     & $60.4$  & $76.5$  & $77.2$  & $64.3$  & $67.5$  & $75.8$  & $69.3$  & $54.2$  & $83.7$  & $75.6$  & $56.7$  & $84.5$  & $70.5$ \\
 SAFN  & $58.9$ & $76.3$ & $81.4$ & $70.4$ & \underline{$73.0$} & $77.8$ & $72.4$ & $55.3$ & $80.4$ & $75.8$ & \underline{$60.4$} & $79.9$ & $71.8$ \\
 DRCN  & $54.0$	& $76.4$	& $83.0$	& $62.1$	& $64.5$	& $71.0$	& $70.8$	& $49.8$	& $80.5$	& $77.5$	& $59.1$	& $79.9$	& $69.0$ \\
 RTNet & \underline{$62.7$}  & $79.3$  & $81.2$  & $65.1$  & $68.4$  & $76.5$  & $70.8$  & $55.3$  & $85.2$  & $76.9$  & $59.1$  & $83.4$  & $72.0$ \\
 RTNet$_{\text{adv}}$ & $\mathbf{63.2}$  & $80.1$  & $80.7$  & $66.7$  & $69.3$  & $77.2$  & $71.6$  & $53.9$  & $84.6$  & $77.4$  & $57.9$  & \underline{$85.5$} & \underline{$72.3$} \\
 BA$^3$US & $60.6$  & \underline{$83.2$} & \underline{$88.4$}  & \underline{$71.8$} & $72.8$  & $\mathbf{83.4}$  & $\mathbf{75.5}$  & $\mathbf{61.6}$ & \underline{$86.5$} & \underline{$79.3$} & $\mathbf{62.8}$  & $\mathbf{86.1}$  & $\mathbf{76.0}$ \\
\hline
  \textbf{\ours (Ours)} & $61.1$  & $\mathbf{84.0}$  & $\mathbf{91.4}$  & $\mathbf{76.5}$  & $\mathbf{75.0}$  & \underline{$81.8$} & \underline{$74.6$} & \underline{$55.6$} & $\mathbf{87.8}$ & $\mathbf{82.3}$  & $57.8$  & $83.5$  & $\mathbf{76.0}$ \\
\hline
\end{tabular}}
\end{center}
\vspace{-5mm}
\caption{\small \textbf{Performance on Office-Home.}  
We highlight the \textbf{best} and \underline{second best} method on each task. 
While the upper section shows results of unsupervised domain adaptation approaches, the lower section shows results of existing partial domain adaptation methods.\ours achieves the best average performance among all compared methods. See supplementary material for standard deviation of each task.
}
\label{table:cls-office-home} 
\vspace{-4mm}
\end{table*}

\setlength{\tabcolsep}{1pt}
\begin{table}[!htbp]

\definecolor{Gray}{gray}{0.90}
\definecolor{LightCyan}{rgb}{0.88,1,1}

\newcolumntype{a}{>{\columncolor{Gray}}c}

\scriptsize
\centering
\begin{adjustbox}{max width=\linewidth}
\begin{tabular}{ l || c  c | a | c c | a  }

\hline
\multicolumn{1}{c}{} & \multicolumn{3}{c}{\textbf{ImageNet-Caltech}} & \multicolumn{3}{c}{\textbf{VisDA-2017}} \\
\hline
 \textbf{Method} &  \textbf{I $\rightarrow$ C} & \textbf{C $\rightarrow$ I} & \textbf{Average} & \textbf{R $\rightarrow$ S} & \textbf{S $\rightarrow$ R} & \textbf{Average} \\
\hline
\hline
 ResNet-50  & $69.7_{\pm 0.8}$ & $71.3_{\pm 0.7}$ & $70.5$ & $64.3$ & $45.3$ &  $54.8$ \\

\hline

 DAN & $71.6$ & $66.5$ & $69.0$ & $68.4$ & $47.6$ & $58.0$ \\
 DANN & $68.7$ & $52.9$ & $60.8$ & $73.8$ & $51.0$ & $62.4$ \\
 ADDA & $71.8_{\pm 0.5}$ & $69.3_{\pm 0.4}$ & $70.6$ & $-$ & $-$ & $-$ \\
 RTN & $72.2$ & $68.3$ & $70.3$ & $72.9$ & $50.0$ & $61.5$ \\
 CDAN+E & $72.5_{\pm 0.1}$ & $72.0_{\pm 0.1}$ & $72.2$ & $-$ & $-$ & $-$ \\

\hline

 PADA  & $75.0_{\pm 0.4}$ & $70.5_{\pm 0.4}$ & $72.8$ & \underline{$76.5$} & $53.5$ & $65.0$ \\
 SAN   & $77.8_{\pm 0.4}$ & $75.3_{\pm 0.4}$ & $76.5$ & $69.7$  & $49.9$ & \\
 IWAN  & $78.1_{\pm 0.4}$ & $73.3_{\pm 0.5}$ & $75.7$ & $71.3$  & $48.6$ & \\
 ETN  & $\mathbf{83.2}_{\pm 0.2}$ & $74.9_{\pm 0.4}$ & \underline{$79.1$} & $-$ & $-$ & $-$ \\
 SAFN & $-$ & $-$ & $-$ & $-$ & \underline{$67.7$}$_{\pm 0.5}$ & $-$ \\
 DRCN    & $75.3$ & \underline{$78.9$} & $77.1$ & $73.2$ & $58.2$ & \underline{$65.7$} \\

\hline
 \textbf{\ours (Ours)} & \underline{$82.3$}$_{\pm 0.1}$ & $\mathbf{81.4}_{\pm 0.6}$ & $\mathbf{81.9}$ & $\mathbf{77.5}_{\pm 0.8}$ & $\mathbf{91.7}_{\pm 0.8}$ & $\mathbf{84.6}$ \\

\hline
\end{tabular}
\end{adjustbox}

\vspace{-2mm}
\caption{\small \textbf{Performance on ImageNet-Caltech and VisDA-2017.}
Our \ours performs the best on both datasets.
}
\label{table:image} 
\vspace{-6mm}
\end{table}

On the challenging Office-Home dataset, our proposed approach obtains very competitive performance, with an average accuracy of $76.0$\% on this dataset (Table~\ref{table:cls-office-home}).
Our method obtains the best on \textbf{6 out of 12} transfer tasks.
Table~\ref{table:image} summarizes the results on ImageNet-Caltech and VisDA-2017 datasets. Our approach once again achieves the best performance, outperforming the next competitive method by a margin of about $\mathbf{2.8}$\textbf{\%} and $\mathbf{18.9}$\textbf{\%} on ImageNet-Caltech and VisDA-2017 datasets respectively. Especially for task S $\rightarrow$ R on VisDA-2017 dataset, our approach significantly outperforms SAFN~\cite{xu2019larger} and DRCN~\cite{li2020deep} by an increase of $\mathbf{24.1}$\textbf{\%} and $\mathbf{33.5}$\textbf{\%} respectively. Note that on the most challenging VisDA-2017 dataset, our approach is still able to distill more positive knowledge from the synthetic to the real domain despite significant domain gap across them. In summary, \ours outperforms the existing PDA methods on all four datasets, showing the effectiveness of our approach in not only identifying the most relevant source classes but also learning more transferable features for partial domain adaptation.

\subsection{Ablation Studies}
We perform the following experiments to test the effectiveness of the proposed modules including the effect of number of target classes on different datasets.

\vspace{1mm}
\noindent\textbf{Effectiveness of Individual Modules.} We conduct experiments to investigate the importance of our three unique modules on three datasets. 
E.g. On Office-Home, as seen from Table~\ref{table:cls-office-home-ablation}, while the Select only module improves the vanilla performance by $8$\%, addition of Label and Mix modules progressively improves the result to obtain the best performance of $76.0$\%. 
This corroborates the fact that both discriminability and invariance of the latent space plays a crucial role in partial domain adaptation in addition to removal of source domain outlier samples. 

\setlength{\tabcolsep}{1pt}
\begin{table*}[!htbp]
\vspace{-3mm}
\definecolor{Gray}{gray}{0.90}
\definecolor{LightCyan}{rgb}{0.88,1,1}

\newcolumntype{a}{>{\columncolor{Gray}}c}
\newcolumntype{P}[1]{>{\centering\arraybackslash}p{#1}}
\newcommand{\cmark}{\ding{51}}%
\newcommand{\xmark}{\ding{55}}%

\scriptsize
\begin{center}
\resizebox{\linewidth}{!}{
\begin{tabular}{  P{0.8cm} | P{0.8cm} | P{0.8cm} || c c c c c c c c c c c c | a  }
\hline
\multicolumn{3}{c}{\textbf{Modules}} & \multicolumn{11}{c}{\textbf{Office-Home}} \\
\hline
\textbf{Select} & \textbf{Label} & \textbf{Mix} &  \textbf{Ar $\rightarrow$ Cl} & \textbf{Ar $\rightarrow$ Pr} & \textbf{Ar $\rightarrow$ Rw} & \textbf{Cl $\rightarrow$ Ar} & \textbf{Cl $\rightarrow$ Pr} & \textbf{Cl $\rightarrow$ Rw} & \textbf{Pr $\rightarrow$ Ar} & \textbf{Pr $\rightarrow$ Cl} & \textbf{Pr $\rightarrow$ Rw} & \textbf{Rw $\rightarrow$ Ar} & \textbf{Pr $\rightarrow$ Cl} & \textbf{Pr $\rightarrow$ Rw} & \textbf{Average} \\
\hline
\hline
\xmark & \xmark & \xmark & $44.2$ & $61.6$ & $75.9$ & $54.6$ & $55.2$ & $65.0$ & $51.0$ & $37.3$ & $69.6$ & $64.8$ & $42.4$ & $71.4$ & $57.7$ \\
\cmark & \xmark & \xmark & $50.6$ & $72.9$ & $79.2$ & $65.4$ & $67.2$ & $71.7$ & $60.8$ & $46.7$ & $77.1$ & $71.9$ & $49.4$ & $77.0$ & $65.8$ \\
\cmark & \cmark & \xmark & $56.1$ & $82.4$ & $89.8$ & $74.2$ & $73.0$ & $81.6$ & $70.8$ & $48.4$ & $87.0$ & $80.1$ & $53.1$ & $81.7$ & $73.2$ \\
\hline
\cmark & \cmark & \cmark  & $61.1$  & $84.0$  & $91.4$  & $76.5$  & $75.0$  & $81.8$ & $74.6$ & $55.6$ & $87.8$ & $82.3$  & $57.8$  & $83.5$  & $76.0$ \\
\hline

\end{tabular}}
\end{center}
\vspace{-5mm}
\caption{\small \textbf{Effectiveness of Different Modules on Office-Home Dataset.} Our proposed approach achieves the best performance with all the modules working jointly for learning discriminative invariant features in partial domain adaptation. 
}
\label{table:cls-office-home-ablation} 
\vspace{-5mm}
\end{table*}

\vspace{0.7mm}
\noindent\textbf{Effectiveness of Discrete Selection.} 
The adverse effects of negative transfer motivated us to adopt a \textit{strong} form of discrete selection of relevant source samples instead of a \textit{weak} form of filtering using soft-weights as adopted in many prior works~\cite{cao2018partialpada, cao2019learning}. In Table~\ref{table:pada-comparison}, we replaced the weighting module in PADA~\cite{cao2018partialpada} with our stronger Select module (PADA w/ SEL) and obtained an average accuracy of $94.9\%$ on Office31 which is $1.6\%$ higher than the original PADA method, showing the superior selection of our Select module. We also adopted the class-weighting ($\gamma$) scheme from PADA~\cite{cao2018partialpada} and used thresholding on top of it (\ours w/ W+T) to replace the Select module in \ours and obtained $94.0\%$ average accuracy. The $4.4\%$ drop shows the ineffectiveness of filtering outliers solely based on the target predictions and the importance of having a dedicated Select module which takes decisions as a function of the source samples. 

\setlength{\tabcolsep}{2pt}
\begin{table}[!tbp]

\definecolor{Gray}{gray}{0.90}
\definecolor{LightCyan}{rgb}{0.88,1,1}

\newcolumntype{a}{>{\columncolor{Gray}}c}
\scriptsize
\begin{center}
\resizebox{\columnwidth}{!}{
\begin{tabular}{ l || c  c   c   c   c   c | a  }
\hline
\multicolumn{8}{c}{\textbf{Office31}} \\
\hline
 \textbf{Method} &  \textbf{A $\rightarrow$ W} & \textbf{D $\rightarrow$ W} & \textbf{W $\rightarrow$ D} & \textbf{A $\rightarrow$ D} & \textbf{D $\rightarrow$ A} & \textbf{W $\rightarrow$ A} & \textbf{Average} \\
\hline
\hline
PADA   & $86.3$  & $99.3$  & $100.0$   & $90.4$  & $91.3$  & $92.6$  & $93.3$ \\
PADA w/ SEL  & $91.8$  & $99.3$  & $96.6$   & $93.8$  & $94.2$  & $93.5$  & $94.9$ \\
\ours w/ W+T & $90.8$  & $99.7$  & $98.7$  & $93.0$  & $91.8$  & $90.2$  & $94.0$   \\
\textbf{\ours} & $99.8$  & $100.0$  & $99.8$  & $98.7$  & $96.1$  & $95.9$  & $98.4$   \\
\hline
\end{tabular}}
\end{center}
\vspace{-6mm}
\caption{\small \textbf{Effectiveness of Discrete Selection.} 
}
\label{table:pada-comparison} 
\vspace{-5mm}
\end{table}

\begin{figure}[!htbp]
\centering
\includegraphics[width=\columnwidth]{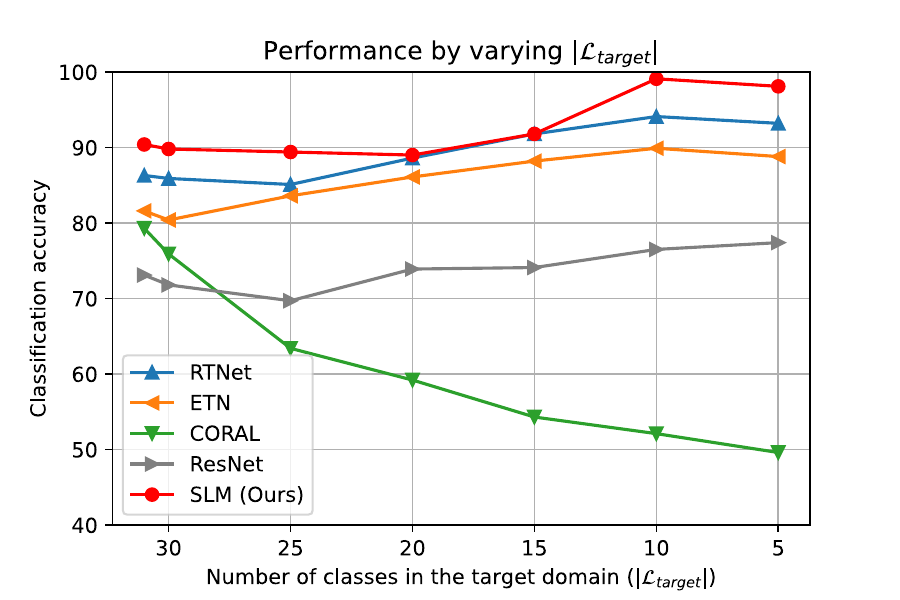} 
\vspace{-6mm}
\caption{\small Performance by varying the number of target classes on A$\rightarrow$W task from Office31 dataset. Best viewed in color.}
\label{fig:var_tgt_cls} 
\vspace{-5mm}
\end{figure}

\vspace{0.7mm}
\noindent\textbf{Comparison with Varying Number of Target Classes.} We compare different methods by varying the target label space. In Figure~\ref{fig:var_tgt_cls}, \ours consistently obtains the best results indicating its advantage in alleviating negative transfer by removing outlier source samples. \ours outperforms all the compared methods even in the case of completely shared space (A31\!$\rightarrow$\!W31), which shows that it does not discard relevant samples incorrectly when there are no outlier classes.

\vspace{0.7mm}
\noindent\textbf{Effectiveness of Different Mixup.} 
We examine the effect of mixup regularization on both domain discriminator and classifier on Office-Home dataset. 
With mixup regularizations working for both discriminator and classifier, the average performance on Office-Home dataset is $76.0$\%. By removing mixup regularization from the training of domain discriminator, it decreases to $73.6$\%. Similarly, by removing mixup regularization from the classifier training, the average performance becomes $73.9$\%. This corroborates the fact that our Mix strategy not only helps to explore intrinsic structures across domains, but also helps to stabilize the domain discriminator. We also explored CutMix~\cite{yun2019cutmix} as an alternative mixing strategy, which can result with new images with information from both the domains are also expected to work well resonating with our motivation. In Table~\ref{table:cutmix}, we replace MixUp with CutMix (\ours w/ CutMix) in the Mix module and obtain an average accuracy of $98.0\%$ on Office31 almost similar to that using MixUp ($98.4\%$). We also tried adding CutMix in addition to MixUp (\ours w/ MixUp+Cutmix) and obtain a similar value of $98.4\%$, with a slight improvement in W$\rightarrow$D task. 

\setlength{\tabcolsep}{2pt}
\begin{table}[!tbp]

\definecolor{Gray}{gray}{0.90}
\definecolor{LightCyan}{rgb}{0.88,1,1}

\newcolumntype{a}{>{\columncolor{Gray}}c}

\scriptsize
\begin{center}
\resizebox{\columnwidth}{!}{
\begin{tabular}{ l || c  c   c   c   c   c | a  }
\hline
\multicolumn{8}{c}{\textbf{Office31}} \\
\hline
 \textbf{Method} &  \textbf{A $\rightarrow$ W} & \textbf{D $\rightarrow$ W} & \textbf{W $\rightarrow$ D} & \textbf{A $\rightarrow$ D} & \textbf{D $\rightarrow$ A} & \textbf{W $\rightarrow$ A} & \textbf{Average} \\
\hline
\hline
\ours w/ CutMix & $99.6$  & $100.0$  & $100.0$  & $97.4$  & $95.7$  & $95.5$  & $98.0$   \\

\ours w/ MixUp & $99.8$  & $100.0$  & $99.8$  & $98.7$  & $96.1$  & $95.9$  & $98.4$   \\

\ours w/ MixUp+CutMix & $99.8$  & $100.0$  & $100.0$  & $98.7$  & $96.1$  & $95.9$  & $98.4$   \\
\hline
\end{tabular}}
\end{center}
\vspace{-6mm}
\caption{\small \textbf{Performance using CutMix.} 
}
\label{table:cutmix} 
\end{table}

\setlength{\tabcolsep}{1pt}
\begin{table}[!htbp]
\definecolor{Gray}{gray}{0.90}
\definecolor{LightCyan}{rgb}{0.88,1,1}

\newcolumntype{a}{>{\columncolor{Gray}}c}
\vspace{-1mm}

\scriptsize
\centering
\begin{adjustbox}{width=0.85\linewidth}
\begin{tabular}{  l || c  c | c  c }

\hline
\textbf{Distance} & \textbf{A $\rightarrow$ D} & \textbf{W $\rightarrow$ A} & \textbf{Cl $\rightarrow$ Pr} & \textbf{Rw $\rightarrow$ Pr} \\
\hline
$\texttt{dist} (\text{S}_{sel}, \text{T})$ & $0.999$ & $0.893$ & $0.819$ & $0.947$\\
$\texttt{dist} (\text{S}_{dis}, \text{T})$ & $1.013$ & $1.144$ & $1.418$ & $1.008$\\
\hline

\end{tabular}
\end{adjustbox}
\vspace{-2mm}
\caption{\small \textbf{Wasserstein Distance between Domains.}  
Table shows values for two randomly sampled tasks from \textbf{Office-31} and \textbf{Office-Home}. The values are normalized by assuming the distance for $\texttt{dist} (\text{S}_{all}, \text{T})$ to be equal to $1.000$, where $\text{S}_{all}$ represents all source samples for the corresponding tasks.}
\label{table:wasser}  

\vspace{-4mm}
\end{table}

\vspace{0.7mm}
\noindent\textbf{Distance between Domains.} Following~\cite{chen2020selective}, we compute the Wasserstein distance between the probability distribution of target samples (T) with that of selected ($\text{S}_{sel}$) and discarded samples ($\text{S}_{dis}$) by the selector network. Table~\ref{table:wasser} shows that $\texttt{dist} (\text{S}_{sel}, \text{T})$ is smaller than $\texttt{dist} (\text{S}_{all}, \text{T})$, while $\texttt{dist} (\text{S}_{dis}, \text{T})$ is greater than $\texttt{dist} (\text{S}_{all}, \text{T})$ on two randomly sampled adaptation tasks from Office31 and Office-Home. The results affirm that samples selected by our selector network are closer to target domain while the discarded samples are very dissimilar to the target domain.

\vspace{1mm}
\noindent\textbf{Additional Ablation Analysis.} We provide additional ablation analyses including effect of Hausdorff distance, soft pseudo-labels, results with different backbones, feature visualizations using t-SNE~\cite{maaten2008visualizing}, etc. in the Appendix.

\vspace{-2mm}

\section{Conclusion}
\label{sec:conclusions}

In this paper, we propose an end-to-end framework for learning discriminative invariant feature representation while preventing negative transfer in partial domain adaptation. While our select module facilitates the identification of relevant source samples for adaptation, the label module enhances the discriminability of the latent space by utilizing pseudo-labels for the target domain samples. The mix module uses mixup regularizations jointly with the other two strategies to enforce domain invariance in latent space.     

\noindent \textbf{Acknowledgement.}
This work is supported by the SERB Grant SRG/2019/001205 and the Defense Advanced Research Projects Agency (DARPA) under Contract No. FA8750-19-C-1001. Any opinions, findings and conclusions or recommendations expressed in this material are those of the author(s) and do not necessarily reflect the views of the Defense Advanced Research Projects Agency.

\clearpage

{\small
\bibliographystyle{ieee_fullname}
\bibliography{egbib}
}

\clearpage

\appendix
\section{Dataset Details}
\label{sec:datasets}
We evaluate the performance of our approach on several benchmark datasets for partial domain adaptation, namely Office31~\cite{saenko2010adapting}, Office-Home~\cite{venkateswara2017deep}, ImageNet-Caltech and VisDA-2017~\cite{peng2017visda}. The following are the detailed descriptions of the above datasets: \\

\begin{figure}[!htbp]
	\begin{center}
	\includegraphics[width=\columnwidth]{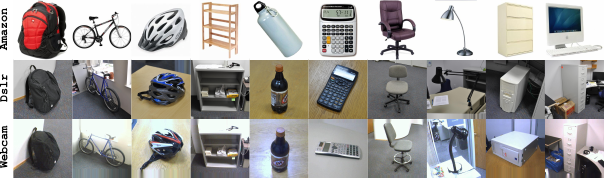}
	\end{center}
	\vskip -0.15in
	\caption{\small \textbf{Sampled Images from Office31 Dataset}. Each row from top to bottom corresponds to the domains Amazon, Dslr and Webcam, respectively. The images in the same column belong to the same class. Best viewed in color.}
	\label{fig:office-31-gallery}
 	\vskip -0.1in
\end{figure}

\vspace{-3mm}
\noindent \textbf{Office31.} This dataset contains 4,110 images distributed among 31 different classes and collected from three different domains: Amazon (A), Webcam (W) and DSLR (D), resulting in 6 transfer tasks. The dataset is imbalanced across domains with 2,817 images belonging to Amazon, 795 images to Webcam, and 498 images to DSLR, making Amazon a larger domain as compared to Webcam and DSLR. For all our experiments, we select the 10 classes shared by Office31 and Caltech256~\cite{griffin2007caltech} as the target categories and obtain the following label spaces: \\
$\mathcal{L}_{source} = \{0, 1, 2, ... , 30\}$.\\
$\mathcal{L}_{target} = \{0, 1, 5, 10, 11, 12, 15, 16, 17, 22\}$.\\
Number of Outlier Classes = $21$.\\
Figure~\ref{fig:office-31-gallery} shows few randomly sampled images from this dataset. The dataset is publicly available to download at:\\ {\small \url{https://people.eecs.berkeley.edu/~jhoffman/domainadapt/#datasets_code}}.

\begin{figure}[!htbp]
	\begin{center}
	\includegraphics[width=\columnwidth]{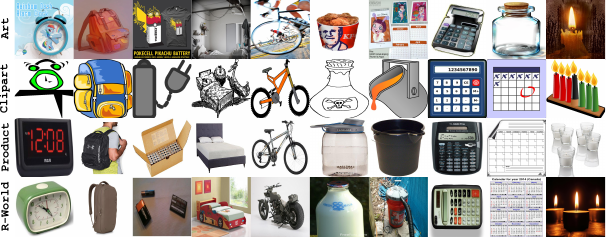}
	\end{center}
	\vskip -0.15in
	\caption{\small \textbf{Sampled Images from Office-Home Dataset}.
	Each row from top to bottom corresponds to the domains Art, Clipart, Product and RealWorld, respectively. The images in the same column belong to the same class. Best viewed in color.}
	\label{fig:office-home-gallery}
 	\vskip -0.1in
\end{figure}

\vspace{1mm}
\noindent \textbf{Office-Home.} This dataset contains 15,588 images distributed among 65 different classes and collected from four different domains: Art (Ar), Clipart (Cl), Product (Pr), and RealWorld (Rw), resulting in 12 transfer tasks. The dataset is split across domains with 2427 images belonging to Art, 4365 images to Clipart, 4439 images to Product, and 4347 images to RealWorld. We select the first 25 categories (in alphabetic order) in each domain as the target classes and obtain the following label spaces:\\
$\mathcal{L}_{source} = \{0, 1, 2, ... , 64\}$.\\
$\mathcal{L}_{target} = \{0, 1, 2, ... , 24\}$.\\
Number of Outlier Classes = $40$.\\
Figure~\ref{fig:office-home-gallery} displays a gallery of sample images for this dataset. The dataset is publicly available to download at: \\ {\small \url{http://hemanthdv.org/OfficeHome-Dataset/}}. \\

\vspace{4mm}

\begin{figure}[!tbp]
	\begin{center}
	\includegraphics[width=\columnwidth]{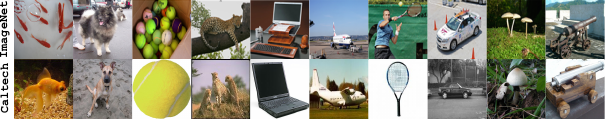}
	\end{center}
	\vskip -0.15in
	\caption{\small \textbf{Sampled Images from ImageNet-Caltech Dataset}.
	The top row corresponds to the ImageNet domain, while the bottom row to the Caltech domain. The images in the same column belong to the same class. Best viewed in color.}
	\label{fig:imagenet-caltech-gallery}
 	\vskip -0.1in
\end{figure}

\vspace{1mm}
\noindent
\textbf{ImageNet-Caltech.} This large-scale dataset consists of two datasets (ImageNet1K~\cite{russakovsky2015imagenet} (I) \& Caltech256~\cite{griffin2007caltech} (C)) as two separate domains and consist of over 14 million images combined. 2 transfer tasks are formed for this dataset. While source domain contains 1,000 and 256 classes for ImageNet and Caltech respectively, each target domain contains only 84 classes that are common across both domains. As it is a general practice to use ImageNet pretrained weights for network initialization, we use the validation set images when using ImageNet as the target domain.
\noindent
Number of Outlier Classes = $172$ for C$\rightarrow$I, $916$ for I$\rightarrow$C.
Figure~\ref{fig:imagenet-caltech-gallery} displays a gallery of sample images for this dataset. The datasets are publicly available to download at:\\
{\small \url{http://www.image-net.org/} \\ \url{http://www.vision.caltech.edu/Image_Datasets/Caltech256/}}.

\begin{figure}[!tbp]
	\begin{center}
	\includegraphics[width=\columnwidth]{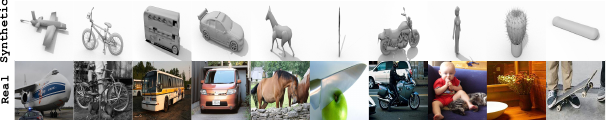}
	\end{center}
	\vskip -0.15in
	\caption{\small \textbf{Sampled Images from VisDA-2017 Dataset}. 
	The top row corresponds to the Synthetic domain, while the bottom row to the Real domain. The images in the same column belong to the same class. Best viewed in color.}
	\label{fig:visda-gallery}
 	\vskip -0.15in
\end{figure}

\vspace{1mm}
\noindent
\textbf{VisDA-2017.} This dataset contains 280,157 images distributed among 12 different classes and two domains. The dataset contains three sets of images: training, validation and testing. The training set contains 152,397 synthetic (S) images, the validation set contains 55,388 real-world (R) images, while the test set contains 72,372 real-world images. For the experiments, the training set is considered as the Synthetic (S) domain, while the validation set as the Real (R) domain, following~\cite{li2020deep}. This results in 2 transfer tasks. The first 6 categories (in alphabetical order) are selected in each of the domains as the target classes, and the following label spaces are obtained: \\
$\mathcal{L}_{source} = \{0, 1, 2, ... , 11\}$.\\
$\mathcal{L}_{target} = \{0, 1, 2, ... , 5\}$.\\
Number of Outlier Classes = $6$.\\
Figure~\ref{fig:visda-gallery} displays a gallery of sample images for this dataset. The dataset is publicly available to download at:\\ {\small \url{http://ai.bu.edu/visda-2017/#download}}.

\section{Implementation Details}
\label{sec:implementation}

The training pipeline pseudo-code for \ours is shown in Algorithm~\ref{alg::pseudocode}.
Following are the detailed description of the implementation we follow for various components of the framework: 

\vspace{1mm}
\noindent
\textbf{Feature Extractor ($\mathcal{G}$).} We use ResNet-50~\cite{he2016deep} backbone for the feature extractor. The overall structure of ResNet-50 is \texttt{Initial Layers, Layer-1, Layer-2, Layer-3, Layer-4, AvgPool, Fc}. The model is initialized with ImageNet~\cite{russakovsky2015imagenet} pretrained weights. Additionally, we add a bottleneck layer of width 256 just after the \texttt{AvgPool} layer to obtain the features and replace all the BatchNorm layers with Domain-Specific Batch-Normalization~\cite{chang2019domain} layers. All the layers till \texttt{Layer-3} are frozen and only the rest of the layers are fine-tuned.

\vspace{1mm}
\noindent\textbf{Selector Network ($\mathcal{H}$).} We use a ResNet-18~\cite{he2016deep} network with the \texttt{Fc} layer replaced with a binary-length fully connected layer as the selector network in our framework. The network is initialized with ImageNet pretrained weights and all the layers are trained while optimization. 

\vspace{1mm}
\noindent\textbf{Classifier ($\mathcal{F}$).} The final \texttt{Fc} layer of ResNet-50 described above is replaced with a task-specific fully-connected layer to form the classifier network of our framework.

\vspace{1mm}
\noindent\textbf{Domain Discriminator ($\mathcal{D}$).} A three-layer fully-connected network is used as the domain discriminator network. It takes the 256-length features obtained from the feature extractor as input. The adversarial training is incorporated using a gradient reversal layer (GRL).

\vspace{1mm}
\noindent\textbf{Hyperparameters.}
All the networks are optimised using mini-batch stochastic gradient descent with a momentum of 0.9. A batch size of 64 is used for Office31 and VisDA-2017 while a batch size of 128 is used for Office-Home and ImageNet-Caltech. For feature extractor an initial learning rate of 5e-5 for the convolutional layers while an initial learning rate of 5e-4 for all the fully-connected layers is used. For the selector network and the domain discriminator an initial learning rate of 5e-3 and 5e-4 are used respectively. The learning rates are decayed following a cosine-annealing strategy as the training progresses. The best models are captured by obtaining the performance on a validation set. We do NOT follow the ten-crop technique~\cite{cao2018partialpada,cao2019learning}, to improve the performance in the inference phase. 
We obtain the best hyperparameters using grid search. All the experiments were averaged over three runs, which used random seed values of $1$, $2$, and $3$ respectively.

\vspace{1mm}
\noindent\textbf{Hardware and Software Details.}
All the experiments were conducted using a single \texttt{NVIDIA Tesla V100-DGXS GPU} with \texttt{32 GigaBytes} of memory, equipped with a \texttt{Intel(R) Xeon(R) CPU E5-2698 v4 @ 2.20GHz}. We used PyTorch v1.4.0, Python v3.6.10 to implement the codes.
\begin{algorithm}[!htb]
\caption{The training pipeline for \ours}
\label{alg::pseudocode}
\textbf{Data:} source data $\mathcal{D}_{source}$ and target data $\mathcal{D}_{target}$.
\\
\textbf{Networks:} Selector Network $\mathcal{H(.)}$, Feature Extractor $\mathcal{G(.)}$, Classifier $\mathcal{F(.)}$, and Domain Discriminator $\mathcal{D(.)}$.
\begin{algorithmic}[1]
\STATE Initialize networks $\mathcal{G(.)}$, $\mathcal{F(.)}$, $\mathcal{H(.)}$, and $\mathcal{D(.)}$ in \ours.
\FOR{$itrn = 1 \to num\_itrn$}
\STATE Obtain the mini-batches $\mathcal{D}_{source}^{b}$ and $\mathcal{D}_{target}^{b}$.
\\\textcolor{magenta}{\# ``Select'' Module}
\STATE Obtain the binary decisions from $\mathcal{H}(\mathcal{D}_{source}^{b})$ \& use them to obtain $\mathcal{D}_{sel}^{b}$ \& $\mathcal{D}_{dis}^{b}$. 
\\\textcolor{magenta}{\# ``Label'' Module}
\STATE Obtain soft pseudo-labels $\hat{y}^{b}$ from $\mathcal{F}(\mathcal{G}(\mathcal{D}_{target}))$ for $\mathcal{D}_{target}$.
\\\textcolor{magenta}{\# ``Mix'' Module}
\STATE Obtain $\mathcal{D}_{\textnormal{\textit{inter\_mix}}}^{b}$, $\mathcal{D}_{\textnormal{\textit{intra\_mix\_s}}}^{b}$, and $\mathcal{D}_{\textnormal{\textit{intra\_mix\_t}}}^{b}$.
\STATE Compute $\mathcal{L}_{sup}$, $\mathcal{L}_{adv}$, $\mathcal{L}_{select}$, $\mathcal{L}_{label}$, and $\mathcal{L}_{mix}$.
\STATE Compute the gradients \& backpropagate for optimization using gradient descent.
\ENDFOR

\end{algorithmic}
\end{algorithm}

\section{Additional Experimental Results}
\label{sec:add_experiments}

\setlength{\tabcolsep}{1pt}
\begin{table*}[!thbp]

\definecolor{Gray}{gray}{0.90}
\definecolor{LightCyan}{rgb}{0.88,1,1}

\newcolumntype{a}{>{\columncolor{Gray}}c}

\scriptsize
\begin{center}
\resizebox{\linewidth}{!}{
\begin{tabular}{ l || c c c c c c c c c c c c | a  }

\hline
\multicolumn{14}{c}{\textbf{Office-Home}} \\
\hline
 \textbf{Method} &  \textbf{Ar $\!\rightarrow\!$ Cl} & \textbf{Ar $\!\rightarrow\!$ Pr} & \textbf{Ar $\!\rightarrow\!$ Rw} & \textbf{Cl $\!\rightarrow\!$ Ar} & \textbf{Cl $\!\rightarrow\!$ Pr} & \textbf{Cl $\!\rightarrow\!$ Rw} & \textbf{Pr $\!\rightarrow\!$ Ar} & \textbf{Pr $\!\rightarrow\!$ Cl} & \textbf{Pr $\!\rightarrow\!$ Rw} & \textbf{Rw $\!\rightarrow\!$ Ar} & \textbf{Rw $\!\rightarrow\!$ Cl} & \textbf{Rw $\!\rightarrow\!$ Pr} & \textbf{Average} \\
\hline
\hline

 ResNet-50   & $47.2_{\pm 0.2}$  & $66.8_{\pm 0.3}$  & $76.9_{\pm 0.5}$  & $57.6_{\pm 0.2}$  & $58.4_{\pm 0.1}$  & $62.5_{\pm 0.3}$  & $59.4_{\pm 0.3}$  & $40.6_{\pm 0.2}$  & $75.9_{\pm 0.3}$  & $65.6_{\pm 0.1}$  & $49.1_{\pm 0.2}$  & $75.8_{\pm 0.4}$  & $61.3$ \\

\hline

DANN     & $43.2_{\pm 0.5}$  & $61.9_{\pm 0.2}$  & $72.1_{\pm 0.4}$  & $52.3_{\pm 0.4}$  & $53.5_{\pm 0.2}$  & $57.9_{\pm 0.1}$  & $47.2_{\pm 0.3}$  & $35.4_{\pm 0.1}$  & $70.1_{\pm 0.3}$  & $61.3_{\pm 0.2}$  & $37.0_{\pm 0.2}$  & $71.7_{\pm 0.3}$  & $55.3$ \\
  CORAL    & $38.2_{\pm 0.1}$  & $55.6_{\pm 0.3}$  & $65.9_{\pm 0.2}$  & $48.4_{\pm 0.4}$  & $52.5_{\pm 0.1}$  & $51.3_{\pm 0.2}$  & $48.9_{\pm 0.3}$  & $32.6_{\pm 0.1}$  & $67.1_{\pm 0.2}$  & $63.8_{\pm 0.4}$  & $35.9_{\pm 0.2}$  & $69.8_{\pm 0.1}$  & $52.5$ \\
 ADDA   &  $45.2$  &  $68.8$  &  $79.2$  &  $64.6$  &  $60.0$  &  $68.3$ &  $57.6$ &  $38.9$ &  $77.5$ &  $70.3$ &  $45.2$ &  $78.3$ &  $62.8$ \\
 RTN & $49.4$  & $64.3$  & $76.2$  & $47.6$  & $51.7$  & $57.7$  & $50.4$  & $41.5$  & $75.5$  & $70.2$  & $51.8$  & $74.8$  & $59.3$ \\
 CDAN+E &  $47.5$  &  $65.9$  &  $75.7$  &  $57.1$  &  $54.1$  &  $63.4$ &  $59.6$ &  $44.3$ &  $72.4$ &  $66.0$ &  $49.9$ &  $72.8$ &  $60.7$ \\
 JDDA     & $45.8_{\pm 0.4}$  & $63.9_{\pm 0.2}$  & $74.1_{\pm 0.3}$  & $51.8_{\pm 0.2}$  & $55.2_{\pm 0.3}$  & $60.3_{\pm 0.2}$  & $53.7_{\pm 0.2}$  & $38.3_{\pm 0.1}$  & $72.6_{\pm 0.2}$  & $62.5_{\pm 0.1}$  & $43.3_{\pm 0.3}$  & $71.3_{\pm 0.1}$  & $57.7$ \\
 SPL & $46.4_{\pm 0.0}$  & $70.5_{\pm 0.6}$  & $77.2_{\pm 0.0}$  & $61.0_{\pm 0.0}$  & $65.2_{\pm 0.0}$  & $73.2_{\pm 0.0}$  & $64.3_{\pm 0.0}$  & $44.7_{\pm 0.0}$  & $79.1_{\pm 0.0}$  & $69.5_{\pm 0.0}$  & $58.0_{\pm 0.0}$  & $79.8_{\pm 0.0}$  & $65.7$ \\
\hline

 PADA    & $53.2_{\pm 0.2}$  & $69.5_{\pm 0.1}$  & $78.6_{\pm 0.1}$  & $61.7_{\pm 0.2}$  & $62.7_{\pm 0.3}$  & $60.9_{\pm 0.1}$  & $56.4_{\pm 0.5}$  & $44.6_{\pm 0.2}$  & $79.3_{\pm 0.1}$  & $74.2_{\pm 0.1}$  & $55.1_{\pm 0.3}$  & $77.4_{\pm 0.2}$  & $64.5$ \\
 SAN   &  $44.4$  &  $68.7$  &  $74.6$  &  $67.5$  &  $65.0$  &  $77.8$ &  $59.8$ &  $44.7$ &  $80.1$ &  $72.2$ &  $50.2$ &  $78.7$ &  $65.3$ \\
 IWAN   &  $53.9$  &  $54.5$  &  $78.1$  &  $61.3$  &  $48.0$  &  $63.3$ &  $54.2$ &  $52.0$ &  $81.3$ &  $76.5$ &  $56.8$ &  $82.9$ &  $63.6$ \\
 ETN     & $60.4_{\pm 0.3}$  & $76.5_{\pm 0.2}$  & $77.2_{\pm 0.3}$  & $64.3_{\pm 0.1}$  & $67.5_{\pm 0.3}$  & $75.8_{\pm 0.2}$  & $69.3_{\pm 0.1}$  & $54.2_{\pm 0.1}$  & $83.7_{\pm 0.2}$  & $75.6_{\pm 0.3}$  & $56.7_{\pm 0.2}$  & $84.5_{\pm 0.3}$  & $70.5$ \\
 SAFN  & $58.9_{\pm 0.5}$ & $76.3_{\pm 0.3}$ & $81.4_{\pm 0.3}$ & $70.4_{\pm 0.5}$ & \underline{$73.0$}$_{\pm 1.4}$ & $77.8_{\pm 0.5}$ & $72.4_{\pm 0.3}$ & $55.3_{\pm 0.5}$ & $80.4_{\pm 0.8}$ & $75.8_{\pm 0.4}$ & \underline{$60.4$}$_{\pm 0.8}$ & $79.9_{\pm 0.2}$ & $71.8$ \\
 DRCN  & $54.0$	& $76.4$	& $83.0$	& $62.1$	& $64.5$	& $71.0$	& $70.8$	& $49.8$	& $80.5$	& $77.5$	& $59.1$	& $79.9$	& $69.0$ \\
 RTNet    & \underline{$62.7$}$_{\pm 0.1}$  & $79.3_{\pm 0.2}$  & $81.2_{\pm 0.1}$  & $65.1_{\pm 0.1}$  & $68.4_{\pm 0.3}$  & $76.5_{\pm 0.1}$  & $70.8_{\pm 0.2}$  & $55.3_{\pm 0.1}$  & $85.2_{\pm 0.3}$  & $76.9_{\pm 0.2}$  & $59.1_{\pm 0.2}$  & $83.4_{\pm 0.3}$  & $72.0$ \\
 RTNet$_{\text{adv}}$ & $\mathbf{63.2}_{\pm 0.1}$  & $80.1_{\pm 0.2}$  & $80.7_{\pm 0.1}$  & $66.7_{\pm 0.1}$  & $69.3_{\pm 0.2}$  & $77.2_{\pm 0.2}$  & $71.6_{\pm 0.3}$  & $53.9_{\pm 0.3}$  & $84.6_{\pm 0.1}$  & $77.4_{\pm 0.2}$  & $57.9_{\pm 0.3}$  & \underline{$85.5$}$_{\pm 0.1}$  & \underline{$72.3$} \\
 BA$^3$US & $60.6_{\pm 0.5}$  & \underline{$83.2$}$_{\pm 0.1}$  & \underline{$88.4$}$_{\pm 0.2}$  & \underline{$71.8$}$_{\pm 0.2}$  & $72.8_{\pm 1.1}$  & $\mathbf{83.4}_{\pm 0.6}$  & $\mathbf{75.5}_{\pm 0.2}$  & $\mathbf{61.6}_{\pm 0.4}$  & \underline{$86.5$}$_{\pm 0.2}$  & \underline{$79.3$}$_{\pm 0.7}$  & $\mathbf{62.8}_{\pm 0.5}$  & $\mathbf{86.1}_{\pm 0.3}$  & $\mathbf{76.0}$ \\
\hline
\end{tabular}}
\end{center}
\vspace{-3mm}
\caption{\small \textbf{Performance on Office-Home.}  
We highlight the \textbf{best} and \underline{second best} method on each task. 
While the upper section shows results of unsupervised domain adaptation approaches, the lower section shows results of existing partial domain adaptation methods.\ours achieves the best average performance among all compared methods.
}
\label{table:cls-office-home-with-var} 
\end{table*}

\setlength{\tabcolsep}{2pt}
\begin{table*}[!thbp]

\definecolor{Gray}{gray}{0.90}
\definecolor{LightCyan}{rgb}{0.88,1,1}

\newcolumntype{a}{>{\columncolor{Gray}}c}
\scriptsize
\begin{center}
\resizebox{\linewidth}{!}{
\begin{tabular}{ l || c  c   c   c   c   c | a  }

\hline
\multicolumn{8}{c}{\textbf{Office31}} \\
\hline
 \textbf{Method} &  \textbf{A $\rightarrow$ W} & \textbf{D $\rightarrow$ W} & \textbf{W $\rightarrow$ D} & \textbf{A $\rightarrow$ D} & \textbf{D $\rightarrow$ A} & \textbf{W $\rightarrow$ A} & \textbf{Average} \\
\hline
\hline

 VGG-16~\cite{Simonyan15} \tiny(ICLR'15)        & $60.3_{\pm0.8}$ & $98.0_{\pm0.6}$ & $99.4_{\pm0.4}$ & $76.4_{\pm0.5}$ & $73.0_{\pm0.6}$ & $79.1_{\pm0.5}$ & $81.0$   \\

\hline

 PADA~\cite{cao2018partialpada} \tiny(ECCV'18)  & \underline{$86.1$}$_{\pm0.4}$ & $\mathbf{100.0}_{\pm0.0}$ & $\mathbf{100.0}_{\pm0.0}$ & $81.7_{\pm0.3}$ & $93.0_{\pm0.2}$ & \underline{$95.3$}$_{\pm0.3}$ & $92.5$   \\
 SAN~\cite{cao2018partialsan} \tiny(CVPR'18)  & $83.4_{\pm0.4}$ & $99.3_{\pm0.5}$ & $\mathbf{100.0}_{\pm0.0}$ & $90.7_{\pm0.2}$ & $87.2_{\pm0.2}$ & $91.9_{\pm0.4}$ & $92.1$   \\
 IWAN~\cite{zhang2018importance} \tiny(CVPR'18)  & $82.9_{\pm0.3}$ & $79.8_{\pm0.3}$ & $88.5_{\pm0.2}$ & \underline{$91.0$}$_{\pm0.3}$ & $89.6_{\pm0.2}$ & $93.4_{\pm0.2}$ & $87.5$   \\
 ETN~\cite{cao2019learning} \tiny(CVPR'19)   & $85.7_{\pm0.2}$ & $\mathbf{100.0}_{\pm0.0}$ & $\mathbf{100.0}_{\pm0.0}$ & $89.4_{\pm0.2}$ & \underline{$95.9$}$_{\pm0.2}$ & $92.3_{\pm0.2}$ & \underline{$93.9$}   \\

\hline
 \textbf{\ours (Ours)} & $\mathbf{92.0}_{\pm0.1}$ & \underline{$99.8$}$_{\pm0.2}$ & \underline{$99.6$}$_{\pm0.5}$ & $\mathbf{98.1}_{\pm0.0}$ & $\mathbf{96.1}_{\pm0.0}$ & $\mathbf{96.0}_{\pm0.1}$ &  $\mathbf{96.9}$  \\
\hline

\end{tabular}}
\end{center}
\vspace{-3mm}
\caption{\small \textbf{Performance on Office31 with VGG-16 backbone.} 
Numbers show the accuracy (\%) of different methods on partial domain adaptation setting. We highlight the \textbf{best} and \underline{second best} method on each transfer task. Our proposed framework, \ours achieves the best performance on 4 out of 6 transfer tasks including the best average performance among all compared methods. 
}
\label{table:cls-office-31-vgg16} \vspace{-1mm}
\end{table*}

\noindent
\textbf{Results on Office-Home Dataset.}
In Table~\ref{table:cls-office-home-with-var}, along with the performance accuracies, we have included the standard deviation for each adaptation task for the Office-Home dataset, as promised in Table-2 of the main paper.

\vspace{2mm}
\noindent
\textbf{Effectiveness on Different Backbone Networks.}
To show that the proposed framework is backbone-agnostic, i.e., it provides the best performance irrespective of the architecture of the feature extractor, we conduct experiments using a VGG-16~\cite{Simonyan15} backbone for the feature extractor. We report the results on the transfer tasks from the Office31 dataset in Table~\ref{table:cls-office-31-vgg16} and compare it with the current state-of-the-art methods. Our method outperforms the previously best results by a margin of $\mathbf{3.0\%}$ on average and achieves new state-of-the-art results. This confirms that our proposed framework for partial domain adaptation is robust with respect to the change of backbone network. 

\vspace{2mm}
\noindent
\textbf{Effectiveness of Individual Modules.}
In Section 4.3 of the main paper, we discussed the importance of the proposed three unique modules on Office-Home dataset. Here, we extend the experiments to Office31 and VisDA-2017 and provide the performance on the transfer tasks in Table~\ref{table:cls-office31-visda17-ablation}. Similar to the results on Office-Home dataset, our approach with all the three modules (Select, Label and Mix) working jointly, works the best on both datasets.

\setlength{\tabcolsep}{1pt}
\begin{table*}[!htbp]

\definecolor{Gray}{gray}{0.90}
\definecolor{LightCyan}{rgb}{0.88,1,1}

\newcolumntype{a}{>{\columncolor{Gray}}c}
\newcolumntype{P}[1]{>{\centering\arraybackslash}p{#1}}
\newcommand{\cmark}{\ding{51}}%
\newcommand{\xmark}{\ding{55}}%

\scriptsize
\begin{center}
\resizebox{\linewidth}{!}{
\begin{tabular}{  P{0.8cm} | P{0.8cm} | P{0.8cm} || c c c c c c | a | c c | a  }

\hline
\multicolumn{3}{c}{\textbf{Modules}} & \multicolumn{7}{c}{\textbf{Office31}} & \multicolumn{3}{c}{\textbf{VisDA-2017}} \\
\hline
\textbf{Select} & \textbf{Label} & \textbf{Mix} &  \textbf{A $\rightarrow$ W} & \textbf{D $\rightarrow$ W} & \textbf{W $\rightarrow$ D} & \textbf{A $\rightarrow$ D} & \textbf{D $\rightarrow$ A} & \textbf{W $\rightarrow$ A} & \textbf{Average} & \textbf{R $\rightarrow$ S} & \textbf{S $\rightarrow$ R} & \textbf{Average} \\
\hline
\hline

\xmark & \xmark & \xmark & $88.0$ & $98.3$ & $95.8$ & $88.8$ & $84.5$ & $80.2$ & $89.3$ & $57.7$ & $56.4$ & $57.0$\\
\cmark & \xmark & \xmark & $91.8$ & $99.3$ & $96.6$ & $93.8$ & $94.2$ & $93.5$ & $94.9$ & $69.0$ & $68.4$ & $68.7$\\
\cmark & \cmark & \xmark & $92.4$ & $99.9$ & $99.2$ & $94.9$ & $95.5$ & $93.8$ & $96.0$ & $77.2$ & $84.8$ & $81.0$\\
\hline
\cmark & \cmark & \cmark & $99.8$ & $100.0$ & $99.8$ & $98.7$ & $96.1$ & $95.9$ & $98.4$ & $77.5$ & $91.7$ & $84.6$\\
\hline

\end{tabular}}
\end{center}
\caption{\small \textbf{Effectiveness of Different Modules on Office31 and VisDA-2017 Datasets.} Our proposed approach achieves the best performance with all the modules working jointly for learning discriminative invariant features in partial domain adaptation.}
\label{table:cls-office31-visda17-ablation} \vspace{1mm}
\end{table*}

\vspace{2mm}
\noindent
\textbf{Effectiveness of Hausdorff Distance.} We investigate the effect of Hausdorff distance (Eqn.~2 in the main paper) in selector network training and find that removing it lowers down performance from $76.0$\% to $73.7$\% on Office-Home dataset, showing its importance in guiding the selector to discard the outlier source samples for effective reduction in negative transfer.
We provide the individual performance of all the transfer tasks on Office-Home dataset in Table~\ref{table:cls-office-home-hausdorff}, which shows that our approach with Hausdorff distance loss works the best in all cases.

\setlength{\tabcolsep}{1pt}
\begin{table*}[!htbp]

\definecolor{Gray}{gray}{0.90}
\definecolor{LightCyan}{rgb}{0.88,1,1}

\newcolumntype{a}{>{\columncolor{Gray}}c}
\newcolumntype{P}[1]{>{\centering\arraybackslash}p{#1}}
\newcommand{\cmark}{\ding{51}}%
\newcommand{\xmark}{\ding{55}}%

\scriptsize
\begin{center}
\resizebox{\linewidth}{!}{
\begin{tabular}{  l || c c c c c c c c c c c c | a  }

\hline
 &  \textbf{Ar $\rightarrow$ Cl} & \textbf{Ar $\rightarrow$ Pr} & \textbf{Ar $\rightarrow$ Rw} & \textbf{Cl $\rightarrow$ Ar} & \textbf{Cl $\rightarrow$ Pr} & \textbf{Cl $\rightarrow$ Rw} & \textbf{Pr $\rightarrow$ Ar} & \textbf{Pr $\rightarrow$ Cl} & \textbf{Pr $\rightarrow$ Rw} & \textbf{Rw $\rightarrow$ Ar} & \textbf{Pr $\rightarrow$ Cl} & \textbf{Pr $\rightarrow$ Rw} & \textbf{Average} \\
\hline
\hline

W/o Hausdorff Loss & $56.2$ &	$83.1$ &	$90.3$ &	$72.6$ &	$71.5$ &	$80.8$ &	$71.4$ &	$51.6$ &	$84.8$ &	$82.5$ &	$57.5$ &	$81.7$ &	$73.7$ \\
\hline
Ours (\ours)   & $61.1$ & $84.0$ & $91.4$ & $76.5$ & $75.0$ & $81.8$ & $74.6$ & $55.6$ & $87.8$ & $82.3$ & $57.8$ & $83.5$ & $76.0$ \\
\hline

\end{tabular}}
\end{center}
\vspace{-3mm}
\caption{\small \textbf{Effectiveness of Hausdorff Triplet Loss on Office-Home Dataset.}  
The table shows the performance of the framework without (top-row) and with (bottom-row) the inclusion of the Hausdorff distance triplet loss. The results highlight the importance of the Hausdorff distance loss in our proposed framework.
}
\label{table:cls-office-home-hausdorff} 
\vspace{-2mm}
\end{table*}

\vspace{1mm}
\noindent
\textbf{Effectiveness of Soft Pseudo-Labels.} We also test the effectiveness of soft pseudo-labels by replacing them with hard pseudo-labels for the target samples and observe that the average performance decreases from $76.0$\% to $72.0$\% on Office-Home dataset. This confirms that soft pseudo-labels are critical in attenuating the unwanted deviations caused by the false and noisy hard pseudo-labels. We provide the performance on each of the transfer tasks from Office-Home in Table~\ref{table:cls-office-home-ablation-hardpl}.

\setlength{\tabcolsep}{1pt}
\begin{table*}[!htbp]

\definecolor{Gray}{gray}{0.90}
\definecolor{LightCyan}{rgb}{0.88,1,1}

\newcolumntype{a}{>{\columncolor{Gray}}c}
\newcolumntype{P}[1]{>{\centering\arraybackslash}p{#1}}
\newcommand{\cmark}{\ding{51}}%
\newcommand{\xmark}{\ding{55}}%

\scriptsize
\begin{center}
\resizebox{\linewidth}{!}{
\begin{tabular}{  l || c c c c c c c c c c c c | a  }

\hline
 &  \textbf{Ar $\rightarrow$ Cl} & \textbf{Ar $\rightarrow$ Pr} & \textbf{Ar $\rightarrow$ Rw} & \textbf{Cl $\rightarrow$ Ar} & \textbf{Cl $\rightarrow$ Pr} & \textbf{Cl $\rightarrow$ Rw} & \textbf{Pr $\rightarrow$ Ar} & \textbf{Pr $\rightarrow$ Cl} & \textbf{Pr $\rightarrow$ Rw} & \textbf{Rw $\rightarrow$ Ar} & \textbf{Pr $\rightarrow$ Cl} & \textbf{Pr $\rightarrow$ Rw} & \textbf{Average} \\
\hline
\hline

W/ Hard Pseudo-labels & $52.5$ &	$79.9$ &	$90.2$ &	$73.5$ &	$72.6$ &	$78.2$ &	$69.9$ &	$47.5$ &	$87.5$ &	$78.6$ &	$50.6$ &	$82.7$ &	$72.0$ \\
\hline
Ours (\ours)  & $61.1$ & $84.0$ & $91.4$ & $76.5$ & $75.0$ & $81.8$ & $74.6$ & $55.6$ & $87.8$ & $82.3$ & $57.8$ & $83.5$ & $76.0$ \\
\hline

\end{tabular}}
\end{center}
\vspace{-1mm}
\caption{\small \textbf{Effectiveness of Soft Pseudo-labels on Office-Home Dataset.} 
Table shows the performance of the framework when we replace the soft pseudo-labels with hard pseudo-labels (top-row) for the target samples. The results justify that the soft pseudo-labels are critical for our framework and attenuate unwanted deviations caused by hard pseudo-labels.
}
\label{table:cls-office-home-ablation-hardpl} 
\vspace{-2mm}
\end{table*}
\vspace{-2mm}

\vspace{1mm}
\noindent
\textbf{Effectiveness of Different MixUp.} 
We examined the effect of mixup regularization on both domain discriminator and classifier separately in Section 4.3 of the main paper. We concluded that our Mix strategy not only helps to explore intrinsic structures across domains, but also helps
to stabilize the domain discriminator. Here, we provide the corresponding performance on each of the transfer tasks of Office-Home in Table~\ref{table:cls-office-home-ablation-mix}.

\setlength{\tabcolsep}{1pt}
\begin{table*}[!htbp]

\definecolor{Gray}{gray}{0.90}
\definecolor{LightCyan}{rgb}{0.88,1,1}

\newcolumntype{a}{>{\columncolor{Gray}}c}
\newcolumntype{P}[1]{>{\centering\arraybackslash}p{#1}}
\newcommand{\cmark}{\ding{51}}%
\newcommand{\xmark}{\ding{55}}%

\scriptsize
\begin{center}
\resizebox{\linewidth}{!}{
\begin{tabular}{  l || c c c c c c c c c c c c | a  }

\hline
 &  \textbf{Ar $\rightarrow$ Cl} & \textbf{Ar $\rightarrow$ Pr} & \textbf{Ar $\rightarrow$ Rw} & \textbf{Cl $\rightarrow$ Ar} & \textbf{Cl $\rightarrow$ Pr} & \textbf{Cl $\rightarrow$ Rw} & \textbf{Pr $\rightarrow$ Ar} & \textbf{Pr $\rightarrow$ Cl} & \textbf{Pr $\rightarrow$ Rw} & \textbf{Rw $\rightarrow$ Ar} & \textbf{Pr $\rightarrow$ Cl} & \textbf{Pr $\rightarrow$ Rw} & \textbf{Average} \\
\hline
\hline

No Domain Discriminator MixUp & $56.2$ &	$81.5$ &	$90.0$ &	$74.0$ &	$71.8$ &	$80.3$ &	$72.2$ &	$50.9$ &	$86.3$ &	$79.8$ &	$58.0$ &	$82.0$ &	$73.6$ \\
No Classifier MixUp & $57.8$ &	$82.9$ &	$88.5$ &	$75.1$ &	$73.6$ &	$79.3$ &	$69.0$ &	$54.9$ &	$86.6$ &	$79.8$ &	$57.6$ &	$81.2$ &	$73.9$ \\
\hline
Ours (\ours)  & $61.1$ & $84.0$ & $91.4$ & $76.5$ & $75.0$ & $81.8$ & $74.6$ & $55.6$ & $87.8$ & $82.3$ & $57.8$ & $83.5$ & $76.0$ \\
\hline

\end{tabular}}
\end{center}
\vspace{-1mm}
\caption{\small \textbf{Effectiveness of Different MixUp on Office-Home Dataset.}  
The table shows the performance of the framework with the exclusion of mixup regularization from the domain discriminator (top-row) and the classsifier (middle-row). The final row shows the results of the proposed \ours framework, which provides the best performance confirming the importance of our Mix strategy. 
}
\label{table:cls-office-home-ablation-mix}
\vspace{-2mm}
\end{table*}

\section{Qualitative Results}
\label{sec:qualitative}

\begin{figure*}[!t]
\captionsetup[subfigure]{labelformat=empty}
\captionsetup[subfigure]{font=scriptsize}
\centering
\subfloat[\textbf{Vanilla}]{
\fbox{\includegraphics[width=0.175\linewidth]{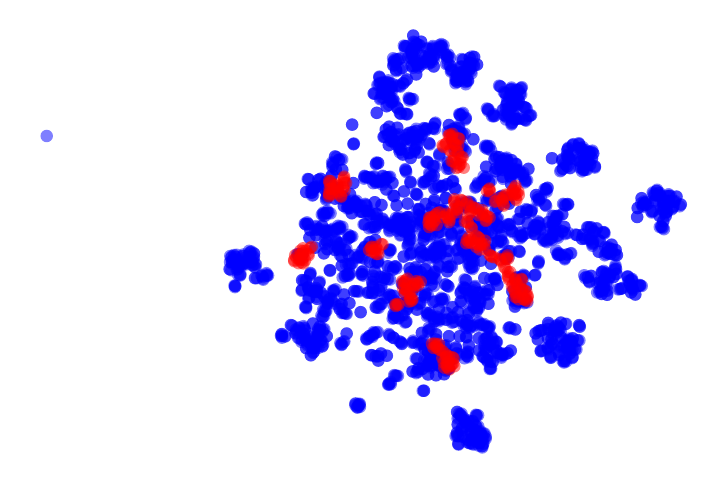}}}
\subfloat[\textbf{Select}]{
\fbox{\includegraphics[width=0.175\linewidth]{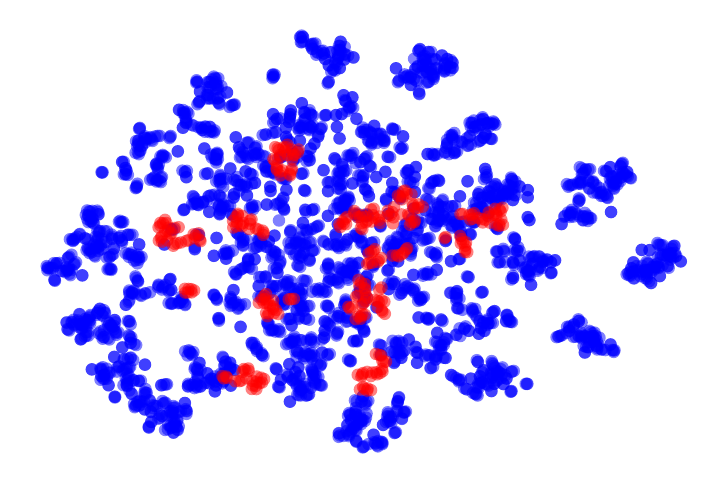}}}
\subfloat[\textbf{Select + Label}]{
\fbox{\includegraphics[width=0.175\linewidth]{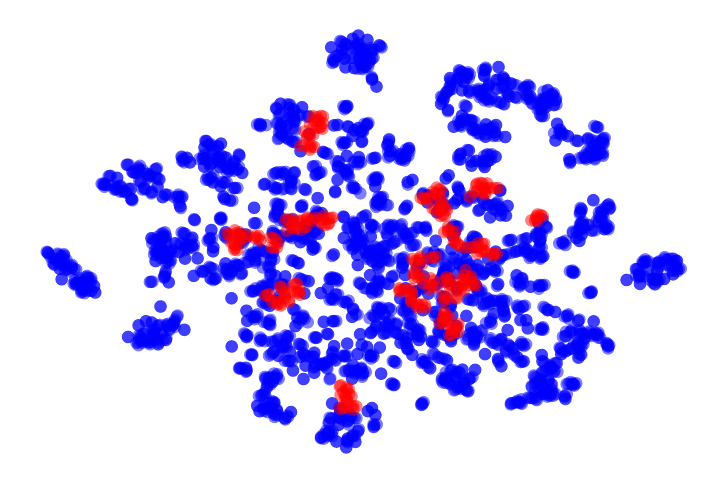}}}
\subfloat[{\centering\textbf{Select + Label + Mix} (\ours)}]{
\fbox{\includegraphics[width=0.175\linewidth]{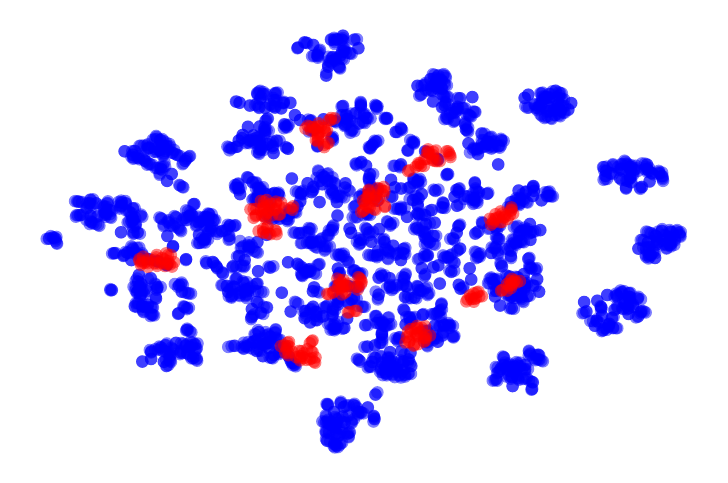}}}

\subfloat[\textbf{Vanilla}]{
\fbox{\includegraphics[width=0.175\linewidth]{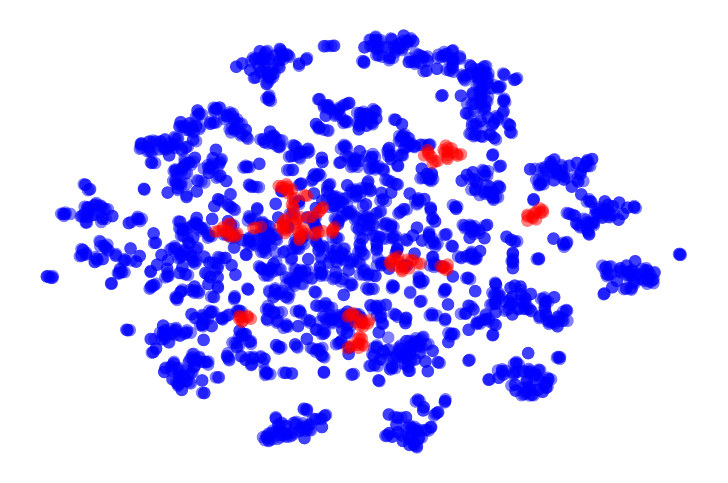}}}
\subfloat[\textbf{Select}]{
\fbox{\includegraphics[width=0.175\linewidth]{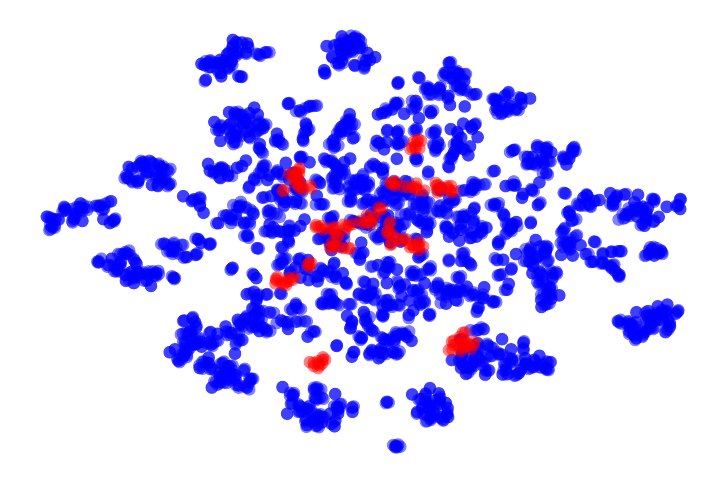}}}
\subfloat[\textbf{Select + Label}]{
\fbox{\includegraphics[width=0.175\linewidth]{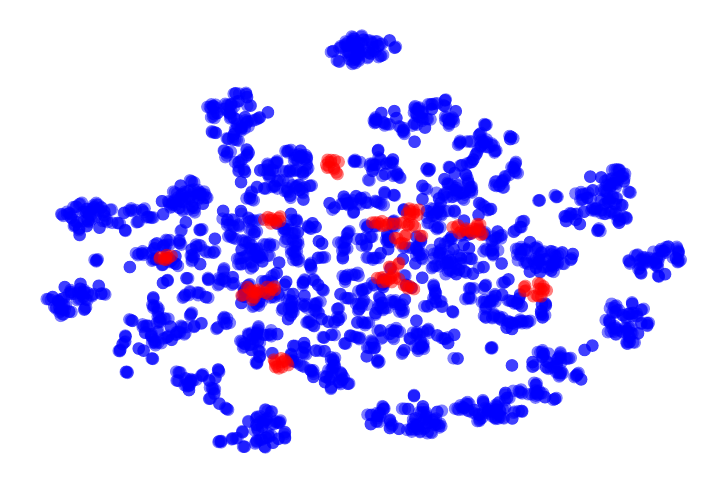}}}
\subfloat[{\centering\textbf{Select + Label + Mix} (\ours)}]{
\fbox{\includegraphics[width=0.175\linewidth]{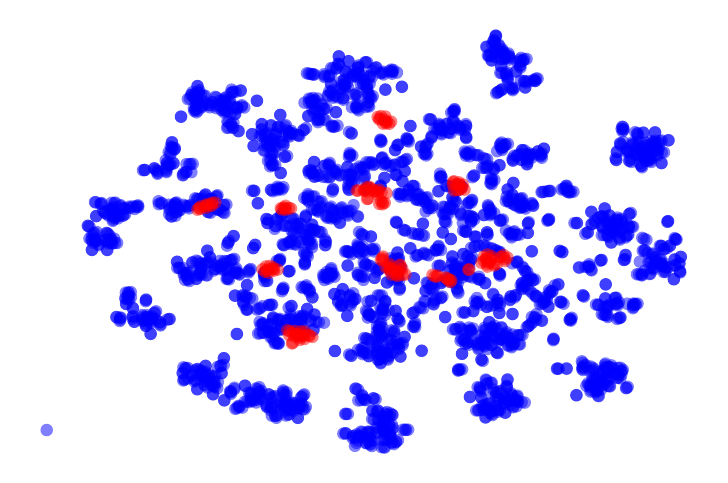}}}

\subfloat[\textbf{Vanilla}]{
\fbox{\includegraphics[width=0.175\linewidth]{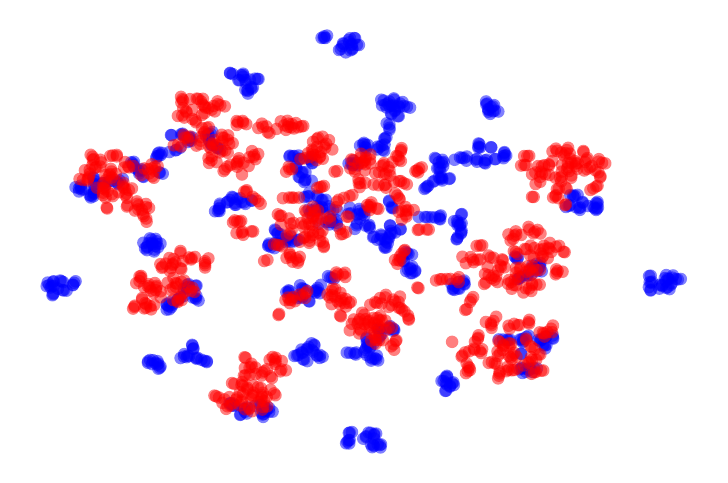}}}
\subfloat[\textbf{Select}]{
\fbox{\includegraphics[width=0.175\linewidth]{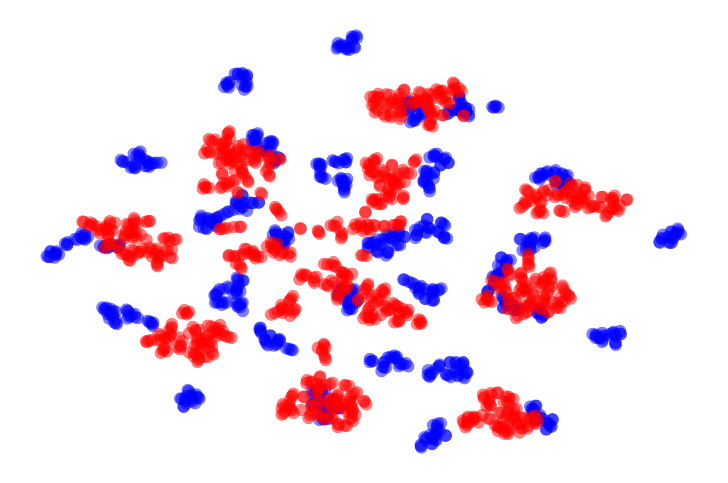}}}
\subfloat[\textbf{Select + Label}]{
\fbox{\includegraphics[width=0.175\linewidth]{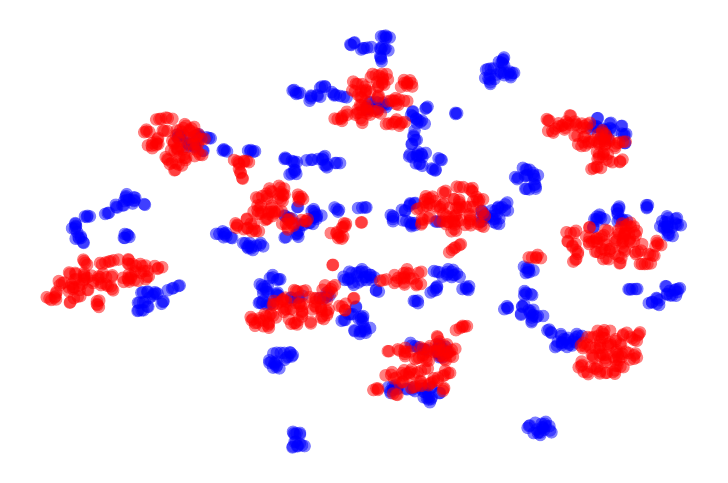}}}
\subfloat[{\centering\textbf{Select + Label + Mix} (\ours)}]{
\fbox{\includegraphics[width=0.175\linewidth]{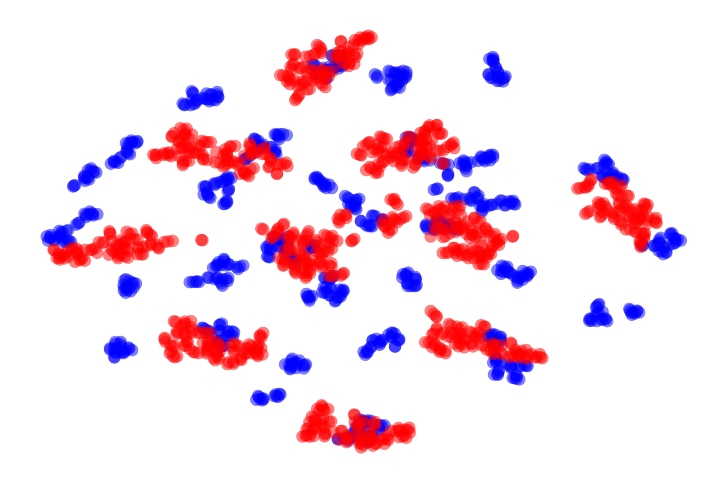}}}

\subfloat[\textbf{Vanilla}]{
\fbox{\includegraphics[width=0.175\linewidth]{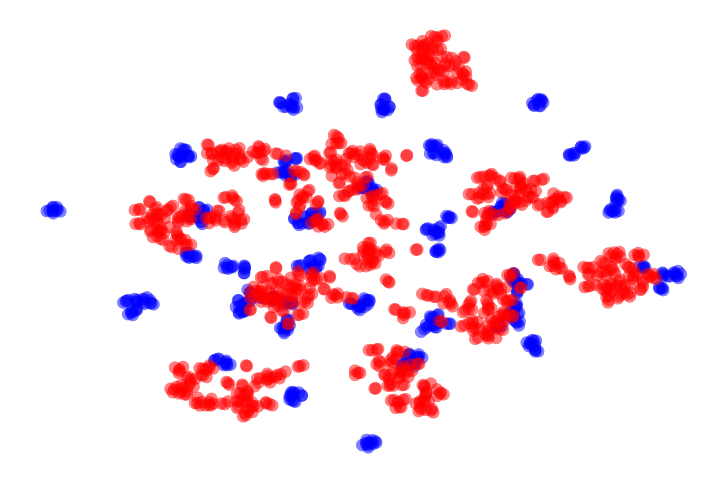}}}
\subfloat[\textbf{Select}]{
\fbox{\includegraphics[width=0.175\linewidth]{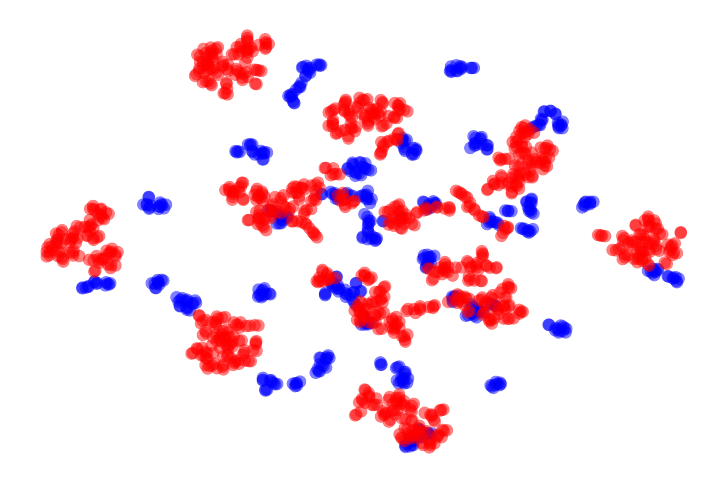}}}
\subfloat[\textbf{Select + Label}]{
\fbox{\includegraphics[width=0.175\linewidth]{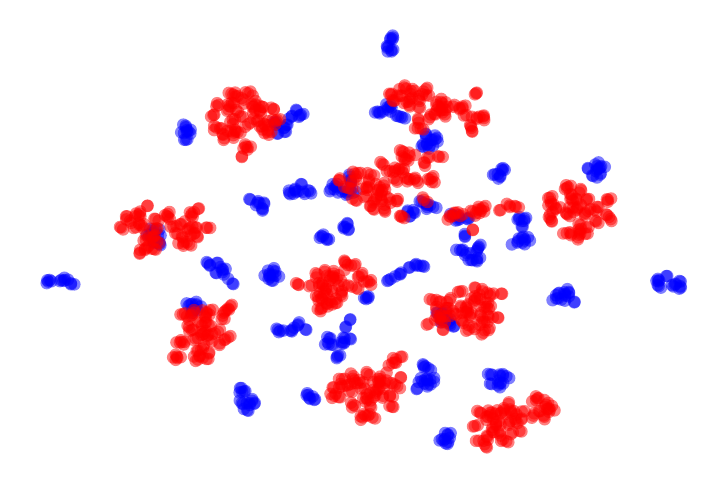}}}
\subfloat[{\centering\textbf{Select + Label + Mix} (\ours)}]{
\fbox{\includegraphics[width=0.175\linewidth]{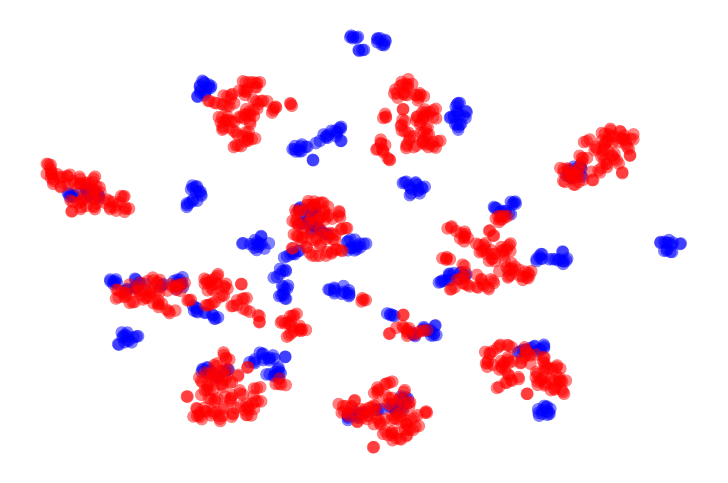}}}
\vspace{-1mm}
\caption{\small \textbf{Feature Visualizations using t-SNE.} Plots show visualization of our approach with different modules on A$\rightarrow$W, A$\rightarrow$D, W$\rightarrow$A, and D$\rightarrow$A tasks repectively (top to down) from Office31 dataset. \textcolor{blue}{Blue} and \textcolor{red}{red} dots represent source and target data respectively.   
As can be seen, features for both target as well as source domain become progressively discriminative and improve from left to right by adoption of our proposed modules. Best viewed in color.}
\label{fig:tsne_supp} 
\vspace{-1mm}
\end{figure*}

\noindent
\textbf{Feature Visualizations.}
We use t-SNE~\cite{maaten2008visualizing} to visualize the features learned using different components of our \ours framework. We choose an UDA setup (similar to DANN~\cite{ganin2016domain}) as a vanilla method and add different modules one-by-one to visualize their individual contribution in learning discriminative features for partial domain adaptation. As seen from Figure~\ref{fig:tsne_supp}, the feature space for vanilla setup lacks dicriminability for both source and target features. The discriminability improves for both source as well as target features as we add ``Select" and ``Label" to the Vanilla setup. The best results are obtained when all three modules ``Select", ``Label" and ``Mix" i.e.,~\ours are added and trained jointly in an end-to-end manner.

\section{Broader Impact and Limitations}
\label{sec:limitations}

\noindent
Our research can help reduce burden of collecting large-scale supervised data in many real-world applications of visual classification by transferring knowledge from models trained on large broad datasets to specific datasets possessing a domain shift. This scenario is quite common as large datasets (e.g. ImageNet~\cite{russakovsky2015imagenet}) can be used for training which contain a broader range of categories while our goal can be to transfer the knowledge to smaller datasets with a smaller number of categories. 
The positive impact that our work could have on society is in making technology more accessible for institutions and individuals that do not have rich resources for annotating newly collected datasets. 
We also believe our approach of selecting relevant source data would motivate the research community to extend it to various open-world problems and would help in training more generalizable models.
Negative impacts of our research are difficult to predict, however, it shares many of the pitfalls associated with standard deep learning models such as susceptibility to adversarial attacks and lack of interpretablity.

\end{document}